\documentclass[10pt,journal]{IEEEtran}  

\usepackage{url}
\usepackage{times}
\usepackage{epsfig}
\usepackage{graphicx}
\usepackage{amsmath}
\usepackage{amssymb}

\usepackage{algorithm}
\usepackage[noend]{algpseudocode}
\usepackage{textcomp} \usepackage{mathtools}
\usepackage{color}

\begin{document}

\title{Skeleton-Based Action Recognition with Multi-Stream Adaptive Graph Convolutional Networks}

\author{Lei~Shi,
        Yifan~Zhang,~\IEEEmembership{Member,~IEEE,}
        Jian~Cheng,~\IEEEmembership{Member,~IEEE,}
        and~Hanqing~Lu,~\IEEEmembership{Senior Member,~IEEE,}
}


\maketitle
\begin{abstract}
    Graph convolutional networks (GCNs), which generalize CNNs to more generic non-Euclidean structures, have achieved remarkable performance for skeleton-based action recognition. However, there still exist several issues in the previous GCN-based models. First, the topology of the graph is set heuristically and fixed over all the model layers and input data.  This may not be suitable for the hierarchy of the GCN model and the diversity of the data in action recognition tasks. Second, the second-order information of the skeleton data, i.e., the length and orientation of the bones, is rarely investigated, which is naturally more informative and discriminative for the human action recognition. In this work, we propose a novel multi-stream attention-enhanced adaptive graph convolutional neural network (MS-AAGCN) for skeleton-based action recognition. The graph topology in our model can be either uniformly or individually learned based on the input data in an end-to-end manner. This data-driven approach increases the flexibility of the model for graph construction and brings more generality to adapt to various data samples. Besides, the proposed adaptive graph convolutional layer is further enhanced by a spatial-temporal-channel attention module, which helps the model pay more attention to important joints, frames and features. Moreover, the information of both the joints and bones, together with their motion information, are simultaneously modeled in a multi-stream framework, which shows notable improvement for the recognition accuracy. Extensive experiments on the two large-scale datasets, NTU-RGBD and Kinetics-Skeleton, demonstrate that the performance of our model exceeds the state-of-the-art with a significant margin.
\end{abstract}

    \begin{IEEEkeywords}
    Skeleton-based action recognition, graph convolutional network, adaptive graph, muti-stream network
    \end{IEEEkeywords}

    \maketitle
    
	\section{Introduction}	
    \label{sec:introduction}
    Action recognition has been widely researched since it plays a significant role in many applications such as video surveillance and human-computer interaction~\cite{wang_action_2013, simonyan_two-stream_2014, tran_learning_2015, carreira_quo_2017, wang_temporal_2018}. 
    Recently, compared with conventional methods that use RGB videos for recognition, the skeleton-based action recognition draws increasingly more attention due to its strong adaptability to the dynamic circumstance and complicated background
    ~\cite{vemulapalli_human_2014,fernando_modeling_2015,du_hierarchical_2015,shahroudy_ntu_2016,liu_spatio-temporal_2016,song_end--end_2017,zhang_view_2017,li_skeleton-based_2018,li_independently_2018,kim_interpretable_2017,ke_new_2017,liu_enhanced_2017,li_skeleton-based_2017,li_skeleton_2017,yan_spatial_2018,tang_deep_2018,zhang_egogesture:_2018}.
    Early deep-learning-based approaches for skeleton-based action recognition manually structure the skeleton data as a sequence of joint-coordinate vectors~\cite{du_hierarchical_2015,shahroudy_ntu_2016,liu_spatio-temporal_2016,song_end--end_2017,zhang_view_2017,li_skeleton-based_2018,li_independently_2018} or as a pseudo-image~\cite{kim_interpretable_2017,ke_new_2017,liu_enhanced_2017,li_skeleton-based_2017,li_skeleton_2017}, which is fed into RNNs or CNNs to generate the prediction.
    However, representing the skeleton data as a vector sequence or a 2D grid cannot fully express the dependencies between correlated joints. 
    The skeleton is naturally structured as a graph in a non-Euclidean space with the joints as vertexes and their natural connections in the human body as edges. 
    Previous approaches cannot exploit the graph structure of the skeleton data and are difficult to generalize to skeletons with arbitrary forms.
    Recently, graph convolutional networks (GCNs), which generalize convolution from image to graph, have been successfully adopted in many applications\cite{kipf_semi-supervised_2016,duvenaud_convolutional_2015,niepert_learning_2016,atwood_diffusion-convolutional_2016,hamilton_inductive_2017,monti_geometric_2017,kipf_neural_2018}. 
    For the skeleton-based action recognition task, Yan et al.~\cite{yan_spatial_2018} first use GCNs to model the skeleton data. 
    They construct a spatial graph based on the natural connections of the joints on the human body and add temporal edges between the corresponding joints in consecutive frames.
    A distance-based sampling function is proposed for constructing the graph convolutional layer, which is then used as a basic block to build the final spatiotemporal graph convolutional network (ST-GCN).

    However, there are three disadvantages for the process of the graph construction in ST-GCN~\cite{yan_spatial_2018}: 
    (1) The skeleton graph used in ST-GCN is heuristically predefined based on the natural connectivity of the human body. Thus it is not guaranteed to be optimal for the action recognition task. 
    For example, the relationship between the two hands is important for recognizing classes such as ``clapping" and ``reading.''
    However, it is difficult for ST-GCN to capture the dependency between the two hands since they are located far away from each other in the predefined human-body-based graphs. 
    (2) The neural networks are hierarchical where different layers contain different levels of semantics. However, the topology of the graph applied in ST-GCN is fixed over all the layers, which lacks the flexibility and capacity to model the multi-level semantics contained in different layers. 
    (3) One fixed graph structure may not be optimal for all the samples of different action classes.
    For classes such as ``wiping face" and ``touching head", the connection between the hands and head should be stronger, but it is not true for some other classes, such as ``jumping up" and ``sitting down". 
    This fact suggests that the graph structure should be data dependent, which, however, is not supported in ST-GCN. 
    
    To solve the above problems, a novel adaptive graph convolutional layer is proposed in this work. It parameterizes two kinds of adaptive graphs for graph convolution.
    One is referred as the global graph, which is obtained by learning the graph adjacency matrix based on the knowledge distilled from the dataset. 
    The learning process uses the task-based loss. Thus the obtained graph topology is more suitable than previous human-body-based graph for the action recognition task.
    Another is referred as the individual graph, whose edges are built according to the feature similarity between graph vertexes. 
    Since the data samples are diverse, the module can capture a unique structure for each input. 
    The two kinds of graphs are fused using a gating mechanism, which can adaptively adjust their importance in each of the model layers. 
    Note that both of the graphs are optimized individually across different layers, thus it can better fit the hierarchical structure of the neural networks. 
    In a word, this data-driven method increases the flexibility of the model for graph construction and brings more generality to adapt to various data samples.

    Besides, since the attention mechanism has been demonstrated the effectiveness and necessity in many tasks~\cite{xu_show_2015, wang_residual_2017, vaswani_attention_2017, hu_gather-excite:_2018}, it is  necessary to investigate it for the skeleton-based action recognition.
    From the spatial perspective, a certain kind of action is usually associated with and characterized by a key subset of the joints. 
    From the temporal perspective, an action flow may contains multiple stages where different sub-stages or frames have different degrees of importance for the final recognition. 
    From the feature perspective, multiple channels of a convolutional feature map contain multiple levels of semantics. Each channel plays different roles for different actions and data samples. 
    These observations inspire us to design a spatial-temporal-channel (STC) attention module to adaptively recalibrate the 
    activations of the joints, frames and channels for different data samples. 
    The module is plugged in each graph convolutional layer, with small amount of parameters yet encouraging performance improvement.  
    
    Another notable problem in previous approaches is that the feature vector attached to each vertex only contains the 2D or 3D coordinates of joints.
    We referred it as the the first-order information of the skeleton data. 
    However, the second-order information, which represents the feature of bones between two joints, is not exploited. 
    Typically, the lengths and directions of bones are naturally more informative and discriminative for action recognition.
    In this work, we formulate the bone information as a vector pointing from its source joint to its target joint.
    Moreover, since the optical flow field have been demonstrated a useful
    modality in the temporal stream for the RGB-based action recognition~\cite{simonyan_two-stream_2014, wang_temporal_2018}, for skeleton data, 
    we propose to extract the coordinate differences of the joints and the bones between two consecutive frames as the motion information to help modeling the temporal evolution of the action.
    Finally, both the joint information and the bone information, together with their motion information, are 
    integrated in a multi-stream framework. 
    All of the streams use the same architecture and the $softmax$ scores of the four streams are fused to generate the final prediction. 
    

    To verify the superiority of the proposed model, namely, the multi-stream attention enhanced adaptive graph convolutional network (MS-AAGCN),  extensive experiments are performed on two large-scale datasets: NTU-RGBD~\cite{shahroudy_ntu_2016} and Kinetics-Skeleton~\cite{kay_kinetics_2017}. 
    Our model achieves the state-of-the-art performance on both of the datasets for skeleton-based action recognition.
    We also visualize the learned adaptive graphs and attention maps of the model and analyze the complementary of the four modalities. 
    
    In addition, since the RGB data contains more abundant appearance information than skeletons, we provide a comparison with the RGB modality and propose a skeleton-guided cropping strategy to fuse the two modalities. 
    The fused model obtains accuracies of 99$\%$ and 96$\%$ in the CV and CS benchmarks of the NTU-RGBD dataset, respectively, which exceeds the other methods with a significant margin.

	 Overall, the main contributions of our work lie in four folds: 
	 (1) An adaptive graph convolutional network is proposed to adaptively learn the topology of the graph in an end-to-end manner, which can better suit the action recognition task, the hierarchical structure of the GCNs and the diverse skeleton samples.
	 (2) A STC-attention module is proposed and embedded in each graph convolutional layer, which can help model learn to selectively focus on discriminative joints, frames and channels.
    (3) The second-order information (bones) of the skeleton data is firstly formulated and combined with the first-order information (joints) in this work, which brings notable improvement for the recognition performance. 
    We further extract the motion information of both joints and bones and integrate these four modalities in a multi-stream framework.
    (4) The effectiveness of the proposed MS-AAGCN is verified on two large-scale datasets for skeleton-based action recognition. Compared with the baseline method, i.e., ST-GCN, it obtains significant improvements of $+7.9\%$ and $+8.5\%$ on the CV and CS benchmarks of the NTU-RGBD dataset, respectively. By combing it with the skeleton-guided cropped RGB data, it obtains additional improvements of $+2.8\%$ and $+6.1\%$. 
    The code is released for future work and to facilitate communication\footnote{https://github.com/lshiwjx/2s-AGCN}.
    
    This paper is an extended version of our previous work~\cite{shi_two-stream_2019} in a number of aspects. 
    First, 
    we optimize the composition scheme of the proposed global and individual graphs
    and introduce a gating mechanism to adaptively adjust the importance of the two graphs.
    Second, we propose a STC-attention module, which turn out to be effective to help the model paying attention to important joints, frames and features. 
    Third, we extend the previous model to a multi-stream framework, which integrates 
    the motion modality for both of the joints and the bones. 
    We provide more extensive experiments and more comprehensive discussion and visualization to demonstrate the effectiveness and the necessity of the proposed modules and data modalities. 
    The model with these designs achieves the accuracy improvements of $+1.1\%$ and $+1.5\%$ on the cross-view and cross-subject benchmarks of the NTU-RGBD dataset for skeleton-based recognition, respectively.
    Finally, we present an effective pose-guided cropping strategy to fuse the skeletons with the RGB modality, which exhibits excellent performance.
    
    The rest of paper is organized as follows. 
    Sec.~\ref{sec:relatedwork} introduces the related work. 
    Sec.~\ref{sec:gcn} formulates the basic GCNs for the skeleton-based action recognition. 
    The components of our proposed MS-AAGCN is introduced in detail in Sec.~\ref{sec:gcn}.
    The ablation study and the comparison with the state-of-the-art methods are shown in Sec.~\ref{sec:experiments}.
    Sec.~\ref{sec:fusion} investigates the fusion of the skeleton data with the RGB modality. 
    Sec.~\ref{sec:visualization} provides some quantitative results and discussions. 
    Sec.~\ref{sec:conclusion} concludes the paper.

	\section{Related work}
    \label{sec:relatedwork}
	\subsection{Skeleton-based action recognition}
	Skeleton-based action recognition has been studied extensively in recent years. 
	Here, we only review the works related to our approach.
    Conventional methods for skeleton-based action recognition usually design handcrafted features to model the human body~\cite{vemulapalli_human_2014,fernando_modeling_2015}. 
    For example, Vemulapalli et al.~\cite{vemulapalli_human_2014} encode the skeletons with their rotations and translations in a Lie group. 
    Fernando et al.~\cite{fernando_modeling_2015} use the rank pooling method to represent the data with the parameters of the ranker.
    
    With the development of the deep learning, data-driven methods have become the mainstream methods, where the most widely used models are RNNs and CNNs. 
    RNN-based methods usually model the skeleton data as a sequence of the coordinate vectors along both the spatial and temporal dimensions, where each of the vectors represents a human body  joint~\cite{du_hierarchical_2015,song_end--end_2017,zhang_view_2017,li_independently_2018, si_skeleton-based_2018}. 
    Du et al.~\cite{du_hierarchical_2015} use a hierarchical bidirectional RNN model to identify the skeleton sequence, which divides the human body into different parts and sends them to different sub-networks. 
    Song et al.~\cite{song_end--end_2017} embed a spatiotemporal attention module in LSTM-based model, so that the network can automatically pay attention to the discriminant spatiotemporal region of the skeleton sequence. 
    Zhang et al.~\cite{zhang_view_2017} introduce the mechanism of view transformation in an LSTM-based model, which automatically translates the skeleton data into a more advantageous angle for action recognition.
    Si et al.~\cite{si_skeleton-based_2018} propose a model with spatial reasoning (SRN) and temporal stack learning (TSL), where the SRN can capture the structural information between different body parts and the TSL can model the detailed temporal dynamics. 
    
    CNN-based methods model the skeleton data as a pseudo-image based on the manually designed transformation rules~\cite{kim_interpretable_2017,liu_enhanced_2017,li_skeleton-based_2017,li_skeleton_2017,cao_skeleton-based_2018}.  
    The CNN-based methods are generally more popular than the RNN-based methods because the CNNs have better parallelizability and are easier for training. 
    Kim et al.~\cite{kim_interpretable_2017} use a one-dimensional residual CNN to identify skeleton sequences where the coordinates of joints are directly concatenated. 
    Liu et al.~\cite{liu_enhanced_2017} propose 10 kinds of spatiotemporal images for skeleton encoding, and enhance these images using visual and motion enhancement methods.
    Li et al.~\cite{li_skeleton_2017} use multi-scale residual networks and various data-augmentation strategies for the skeleton-based action recognition.
    Cao et al.~\cite{cao_skeleton-based_2018} design a permutation network to learn an optimized order for the rearrangement of the joints. 
 
    However, both the RNNs and CNNs fail to fully represent the structure of the skeleton data because the skeleton data are naturally embedded in the form of graphs rather than a vector sequence or a 2D grid. 
    Recently, Yan et al.~\cite{yan_spatial_2018} propose a spatiotemporal graph convolutional network (ST-GCN) to directly model  the skeleton data as the graph structure. 
    It eliminates the requirement for designing handcrafted transformation rules to transform the skeleton data into vector sequences or pseudo-images, thus achieves better performance. 
    Based on this, Tang et al.~\cite{tang_deep_2018} further propose a selection strategy of the key frames with the help of reinforcement learning.
    
    \subsection{Graph convolutional neural networks}
    The input of traditional CNNs is usually low-dimensional regular grids, such as image, video and audio.
    However, it is not straightforward to use the CNN to model the graph data, which always has arbitrary size and shape. 
    The graph is more common and general in the wild, such as social network, molecule and parse tree. 
    How to operate on graphs has been explored extensively over decade and now the most popular solution is to use the graph convolutional networks (GCNs). 
    The GCNs is similar with the traditional CNNs, but it can generalize the convolution from image to graph of arbitrary size and shape~\cite{shuman_emerging_2013,bruna_spectral_2014,henaff_deep_2015,niepert_learning_2016,atwood_diffusion-convolutional_2016,kipf_semi-supervised_2016,duvenaud_convolutional_2015,defferrard_convolutional_2016,hamilton_inductive_2017,monti_geometric_2017,kipf_neural_2018}. 
    
    The principle of constructing GCNs mainly follows two streams: spatial perspective and spectral perspective.
    Spatial perspective methods directly perform convolution on the graph vertexes and their neighbors. 
    The key lies in how to construct the locally connected neighborhoods from the graph which misses the implicit order of vertexes and edges. 
    These methods always extracts the neighbors based on the manually designed rules~\cite{duvenaud_convolutional_2015,niepert_learning_2016,hamilton_inductive_2017,monti_geometric_2017,kipf_neural_2018, wang_videos_2018}.
    Niepert et al.~\cite{niepert_learning_2016} sample the neighborhoods for each of the vertexes based on their distances in the graph. A normalization algorithm is proposed to crop excess vertexes and pad dummy vertexes. 
    Wang and Gupta.~\cite{wang_videos_2018} represent the video as a graph containing persons and detected objects for action recognition. The neighborhood of each vertex is defined according to the feature similarity and the spatial-temporal relations. 
    
    In contrast to the spatial perspective methods, spectral perspective methods use the eigenvalues and eigen vectors of the graph Laplace matrices. 
    These methods perform graph convolution in the frequency domain with the help of the graph Fourier transform~\cite{shuman_emerging_2013}, which does not need to extract locally connected regions from graphs at each convolution step~\cite{bruna_spectral_2014,henaff_deep_2015,kipf_semi-supervised_2016,defferrard_convolutional_2016}. 
    Defferrard et al.~\cite{defferrard_convolutional_2016} propose to use recurrent Chebyshev polynomials as the filtering scheme which is more efficient than previous polynomial filter. 
    Kipf and Welling~\cite{kipf_semi-supervised_2016} further simplified this approach using the first-order approximation of the spectral graph convolutions. 
    This work follows the spatial perspective methods. 

    \section{Graph Convolutional Networks}
    \label{sec:gcn}
    In this section, we introduce a 
    basic graph convolutional network and its implementation for skeleton-based action recognition. 
    
    \subsection{Graph construction}
\begin{figure}[!htb]
	\centering
	\includegraphics[width=\linewidth]{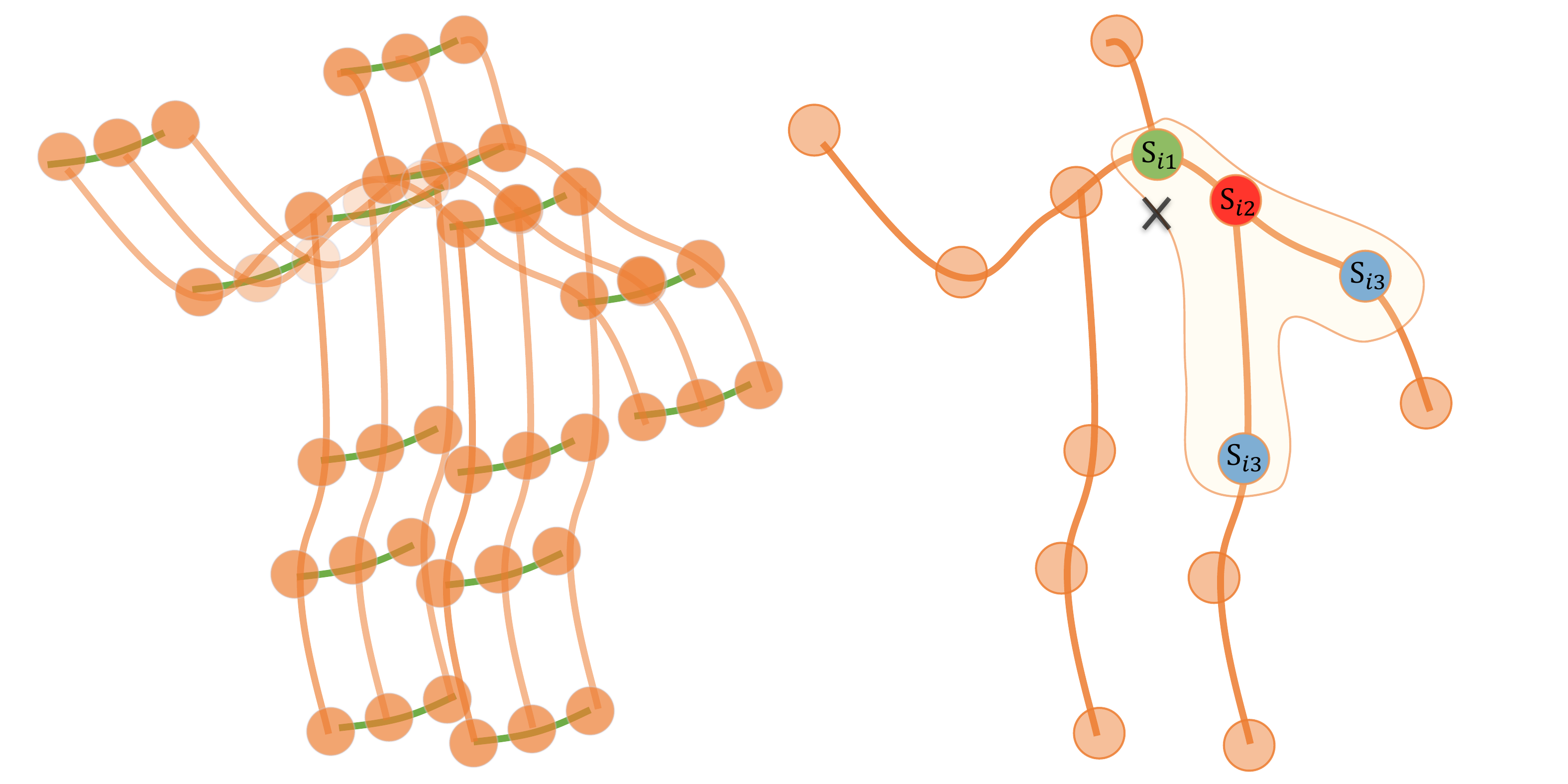}
	\caption{Left: Illustration of the spatiotemporal graph. Right:  Illustration of the mapping strategy. Different colors denote different subsets.}
	\label{fig:skeleton}
	\end{figure}

    The raw skeleton data in one frame are always represented by a sequence of vectors.
    Each vector represents the 2D or 3D coordinates of the corresponding human joint. 
    A complete action contains multiple frames with different lengths for different samples.
    We use a spatiotemporal graph to model the structured information among these joints along both the spatial and temporal dimensions.
    Here, the spatial dimension refers to the joints in the same frame, and the temporal dimension refers to the same joints over all of the frames. 
    The left 
    sub-figure in Fig.~\ref{fig:skeleton} presents an example of the constructed spatiotemporal skeleton graph, where the joints are represented as the vertexes and their natural connections in the human body are represented as the spatial edges (the orange lines in Fig.~\ref{fig:skeleton}, left). 
    For the temporal dimension, the corresponding joints in two consecutive frames are connected with temporal edges the green lines in Fig.~\ref{fig:skeleton}, left).
    The coordinate vector of each joint is set as the attribute of the corresponding vertex. 
    Since the graph is intrinsic and is built based on the natural connectivity of the human body, we refer it as the human-body-based graph. 
    
    \subsection{Graph convolution}
    \label{sec:graphconvolution}

    Given the graph defined above, for the spatial dimension, the graph convolution operation on vertex $v_i$ is formulated as:
 	\begin{equation}
    \label{eq:stgcn}
    \mathit{f}_{out}(v_{i})=\sum_{v_{j}\in \mathcal{B}_i} \frac{1}{Z_{ij} } \mathit{f}_{in}(v_{j})\cdot{}w(l_{i}(v_{j}))
    \end{equation}
    where $\mathit{f}$ denotes the feature map and $v$ denotes the vertex of the graph. 
    $\mathcal{B}_i$ denotes the sampling area of the convolution for $v_i$, which is defined as the 1-distance 
    neighboring vertexes ($v_j$) of the target vertex ($v_i$). 
    $w$ is the weighting function similar to the traditional 
    convolution operation, which provides a weight vector based on the given input. 
    Note that the number of weight vectors of convolution is fixed, while the number of vertexes in $\mathcal{B}_i$ is varied. 
    So a mapping function $l_{i}$ is required to map all neighboring vertexes into a fix-numbered subsets each of which is associated with a unique weight vector. 
    The right 
    sub-figure in Fig.~\ref{fig:skeleton} shows this mapping strategy, where $\times$ represents the center of gravity of the skeleton. $\mathcal{B}_i$ is the sampling area enclosed by the curve.
    In detail, the strategy empirically sets the kernel size as 3 and naturally divides $\mathcal{B}_i$ into 3 subsets: 
    $\mathcal{S}_{i1}$ is the vertex itself (the red circle in Fig.~\ref{fig:skeleton}, right); 
    $\mathcal{S}_{i2}$ is the centripetal subset, which contains the neighboring vertexes that are closer to the center of gravity (the green 
    dot); 
    $\mathcal{S}_{i3}$ is the centrifugal subset, which contains the neighboring vertexes that are farther from the center of gravity (the blue 
    dot). 
    $Z_{ij}$ denotes the cardinality of $\mathcal{S}_{ik}$ that contains $v_j$. It aims to balance the contribution of each subset.

    \subsection{Implementation}
    The implementation of the graph convolution for the spatial dimension is not straightforward.
    Concretely, the feature map of the network is actually a tensor $\mathit{f}\in \mathbb{R}^{C\times T\times N}$, where $N$ denotes the number of  vertexes, $T$ denotes the temporal length and $C$ denotes the number of channels. 
    To implement the GCN in code, Eq.~\ref{eq:stgcn} is transformed into
    \begin{equation}
    \label{eq:stgcni}
    \mathbf{f}_{out} = \sum_k^{K_v} \mathbf{W}_k (\mathbf{f}_{in} {\mathbf{A}}_k)
    \end{equation}
where $K_v$ denotes the kernel size of the spatial dimension. With the mapping strategy designed above, $K_v$ is set to $3$. 
    $\mathbf{A}_k = {\mathbf{\Lambda}}_k^{-\frac{1}{2}} {\mathbf{\bar{A}}}_k {\mathbf{\Lambda}}_{k}^{-\frac{1}{2}}$, where $\mathbf{\bar{A}}_k \in \mathbb{R}^{N\times N}$ is similar to the adjacency matrix. Its element $\mathbf{\bar{A}}_k^{ij}$ indicates whether the vertex $v_j$ is in the subset $S_{ik}$ of the vertex $v_i$. 
    It is used to 
    select the connected vertexes in a particular subset from $\mathbf{f}_{in}$ for the corresponding weight vector.
    $\mathbf{\Lambda}_k^{ii}=\sum_j (\mathbf{\bar{A}}_k^{ij})+\alpha$ is the normalized diagonal matrix. 
    $\alpha$ is set to $0.001$ to avoid empty rows. 
    $\mathbf{W}_k \in \mathbb{R}^{C_{out}\times C_{in}\times 1 \times 1}$ is the weight vector of the $1\times 1$ convolution operation, which represents the weighting function $w$ in Eq.~\ref{eq:stgcn}. 
    
    For the temporal dimension, since the number of neighbors for each vertex is fixed as $2$ (corresponding joints in the two adjacent frames), it is straightforward to perform the graph convolution similar to the classical convolution operation. Concretely, we perform a $K_t\times 1$ convolution on the output feature map calculated above, where $K_t$ is the kernel size of the temporal dimension.

    With above formulations, multiple layers of spatiotemporal graph convolution operations are applied on the graph to extract the high-level features. 
    The global average pooling layer and the $softmax$ classifier are then used to predict the action categories based on the extracted features.

    \section{Multi-stream attention-enhanced adaptive graph convolutional network}
    \label{sec:msaagcn}
    In this section, we introduce the components of the proposed multi-stream attention-enhanced adaptive graph convolutional network (MS-AAGCN) in detail. 
	\subsection{Adaptive graph convolutional layer}
    \label{sec:agcn}
	The spatiotemporal graph convolution for the skeleton data described above is calculated based on a intrinsic human-body-based graph, which may not be the best choice as explained in Sec.~\ref{sec:introduction}.
    To solve this problem, we propose an adaptive graph convolutional layer. 
    It makes the topology of the graph optimized together with other parameters of the network in an end-to-end learning manner. 
    The graph is unique for different layers and samples, which greatly increases the flexibility of the model. 
    Meanwhile, it is designed as a residual branch, which guarantees the stability of the original model.

    In detail, according to the Eq.~\ref{eq:stgcni}, the graph is actually determined by the adjacency matrix and the mask, i.e., $\mathbf{A}_k$ and  $\mathbf{M}_k$, respectively. 
    $\mathbf{A}_k$ determines whether there are connections between two vertexes and $\mathbf{M}_k$ determines the strength of the connection. 
    To make the graph topology adaptive, we modify the Eq.~\ref{eq:stgcni} into the following form:
    \begin{equation}
    \label{eq:nlstgcn}
    \mathbf{f}_{out} = \sum_k^{K_v} \mathbf{W}_k \mathbf{f}_{in} ({\mathbf{B}_{k}}+\alpha{\mathbf{C}_{k}})
    \end{equation}
    The main difference lies in the adjacency matrix of the graph, which is divided into two sub-graphs: $\mathbf{B}_k$ and $\mathbf{C}_k$.
    
    \textbf{The first sub-graph ($\mathbf{B}_k$)} is the global graph learned from the data. It represents the graph topology that is more suitable for the action recognition task. 
    It is initialized with the adjacency matrix of the human-body-based graph, i.e., $\mathbf{A}_k$ in Eq.~\ref{eq:stgcni}. 
    Different with $\mathbf{A}_k$, the elements of $\mathbf{B}_k$ are parameterized and updated together with other parameters in the training process.
	There are no constraints on the value of $\mathbf{B}_k$, which means that the graph is completely learned according to the training data. 
	With this data-driven method, the model can learn graphs that are fully targeted to the recognition task.
	$\mathbf{B}_k$ is unique for each layer, thus is more individualized for different levels of semantics contained in different layers.

    \textbf{The second sub-graph} ($\mathbf{C}_{k}$) is the individual graph which learns a unique topology for each sample. 
    To determine whether there is a connection between two vertexes and how strong is the connection, we use the normalized embedded Gaussian function to estimate the feature similarity of the two vertexes as:
    \begin{equation}
    \label{eq:gaussian}
    \mathit{f}(v_i,v_j)=\frac{e^{\theta(v_i)^T\phi(v_j)}}{\sum_{j=1}^N{e^{\theta(v_i)^T\phi(v_j)}}}
    \end{equation}
    where $N$ is the number of vertexes. We use the dot product to measure the similarity of the two vertexes in an embedding space.
    In detail, given the input feature map $\mathit{f_{in}} \in \mathbb{R}^{C_{in}\times T \times N}$, we first embed it into the embedding space $\mathbb{R}^{C_{e}\times T \times N}$ with two embedding functions, i.e., $\theta$ and $\phi$. 
    Here, through extensive experiments, we choose the $1\times 1$ convolutional layer as the embedding function. 
    The two embedded feature maps are reshaped to matrix $\mathbf{M}_{\theta k}\in\mathbb{R}^{N\times C_{e}T}$ and matrix $\mathbf{M}_{\phi k}\in\mathbb{R}^{C_{e}T \times N}$. 
    They are then multiplied to obtain a similarity matrix $\mathbf{C}_k\in \mathbb{R}^{N\times N}$, whose element $\mathbf{C}_k^{ij}$ represents the similarity of the vertex $v_i$ and the vertex $v_j$. The value of the matrix is normalized to $0-1$, which is used as the soft edge of the two vertexes.
    Since the normalized Gaussian is equipped with a $softmax$ operation, we can calculate $\mathbf{C}_k$ based on Eq.\ref{eq:gaussian} as follows:
    \begin{equation}
    \mathbf{C}_k = SoftMax(\mathbf{f_{in}}^T\mathbf{W}^T_{\theta k}\mathbf{W}_{\phi k} \mathbf{f_{in}})
    \end{equation}
    where $\mathbf{W_\theta}\in\mathbb{R}^{C_{e}\times C_{in} \times 1 \times 1}$ and $\mathbf{W_\phi}\in\mathbb{R}^{C_{e}\times C_{in} \times 1 \times 1}$ are the parameters of the embedding functions $\theta$ and $\phi$, respectively. 
    
    \begin{figure}[tb]
	\centering
	\includegraphics[width=0.95\linewidth]{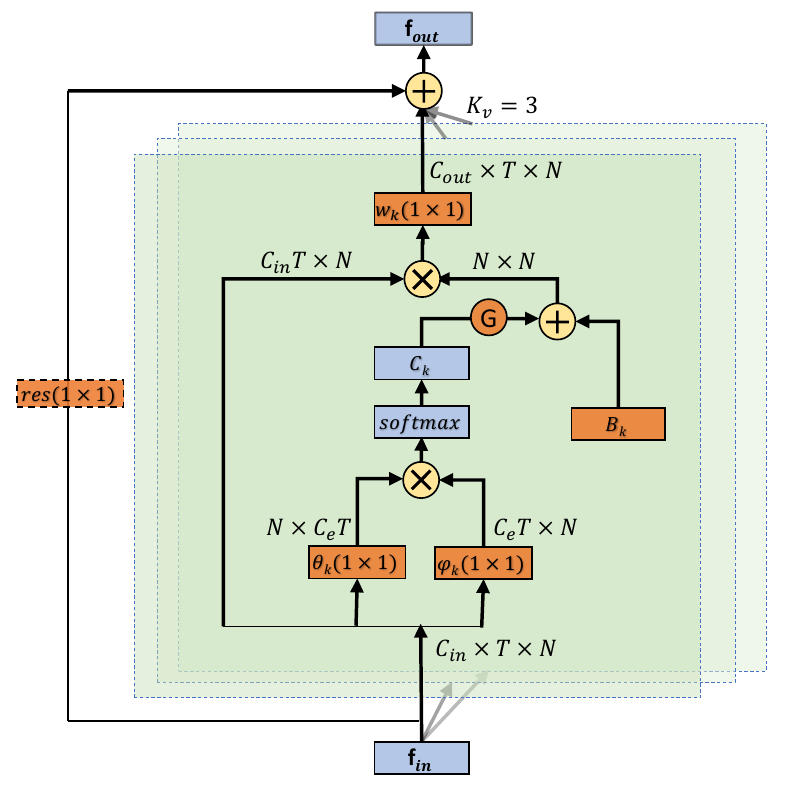}
	\caption{Illustration of the adaptive graph convolutional layer (AGCL). There are two kinds of graphs in each layer, i.e., $\mathbf{B}_k$ and $\mathbf{C}_k$. The orange box indicates that the part is the parameter of the network and is updated during the training process. $\theta$ and $\phi$ are two embedding functions whose kernel size is $(1\times 1)$. $K_v$ denotes the number of subsets. $\oplus$ denotes the element-wise addition. $\otimes$ denotes the matirx multiplication. $G$ is the gate that controls the importance of the two kinds of graphs. The residual box (dotted line) is only needed when $C_{in}$ is not the same as $C_{out}$.}
	\label{fig:nlgcn}	
	\end{figure}
    
    \textbf{Gating mechanism:}
    The global graph determines the basic graph topology for the action recognition task and the individual graph 
    adds 
    individuality according to the various sample features. 
    In the experiments we found that the individual graph is required more strongly in the 
    top layers than the 
    bottom layers. 
    It is reasonable since the receptive field of the 
    bottom layer is smaller, which limits the ability of learning the graph topology from diverse samples.
    Furthermore, the information contained in the top layers is more semantic, which is more variable and requires more 
    individuality of the graph topology. 
    The individual graph is easier to meet the requirement because it is constructed based on the input features and is individual for each of the samples.
    Based on these observations, we use a gating mechanism to adjust the importance of the individual graph for different layers. 
    In detail, the $\mathbf{C}_k$ is multiplied with a parameterized coefficient $\alpha$ that is unique for each layer and is learned and updated in the training process. 
    
    \textbf{Initialization:} 
    \label{sec:init}
    In the experiments we found that the graph topology changes dramatically in the 
    early stage of the training process, thus causes 
    instable and affect the convergence of the model. 
    To stabilize the training, we tried two strategies.
    The first strategy is using $\mathbf{A}_k + \alpha\mathbf{B}_k + \beta\mathbf{C}_k$ as the adjacency matrix where $\mathbf{A}_k$ is the human-body-based graph which is fixed. 
    The $\mathbf{B}_k$, $\mathbf{C}_k$, $\alpha$ and $\beta$ are initialized to be $0$, thus the $\mathbf{A}_k$ will dominate the early stage of the training.
    The second strategy is initializing the $\mathbf{B}_k$ with $\mathbf{A}_k$ and blocking the propagation of the gradient for $\mathbf{B}_k$ at the early stage of the training process until the training stabilizes. 
    The second strategy is verified slightly better in Sec.~\ref{sec:ablation}. 
    The weights of the embedding functions ($\theta$ and $\phi$) and the fusion coefficient $\alpha$ are all initialized as $0$.
    
    The overall architecture of the adaptive graph convolutional layer (AGCL) is 
    shown in Fig.~\ref{fig:nlgcn}. 
    The kernel size of the graph convolution ($K_v$) is set to $3$. 
    $w_k$ is the weighting function introduced in Eq.~\ref{eq:stgcn}, whose parameter is $\mathbf{W}_k$ introduced in Eq.~\ref{eq:nlstgcn}. 
    A residual connection, similar to~\cite{he_deep_2016}, is added for each layer, which allows the layer to be inserted into any existing models without breaking its initial behavior. If the number of input channels is different from the number of output channels, a $1\times 1$ convolution (orange box with dashed line in Fig.~\ref{fig:nlgcn}) will be inserted in the residual path to transform the input to match the output in the channel dimension. 
    $G$ is  the  gate  that  controls the  semantics  of  the  two  kinds  of  graphs.
   
    \subsection{Attention module}
    \label{sec:attention}
    
\begin{figure*}[tb]
	\centering
	\includegraphics[width=0.7\linewidth]{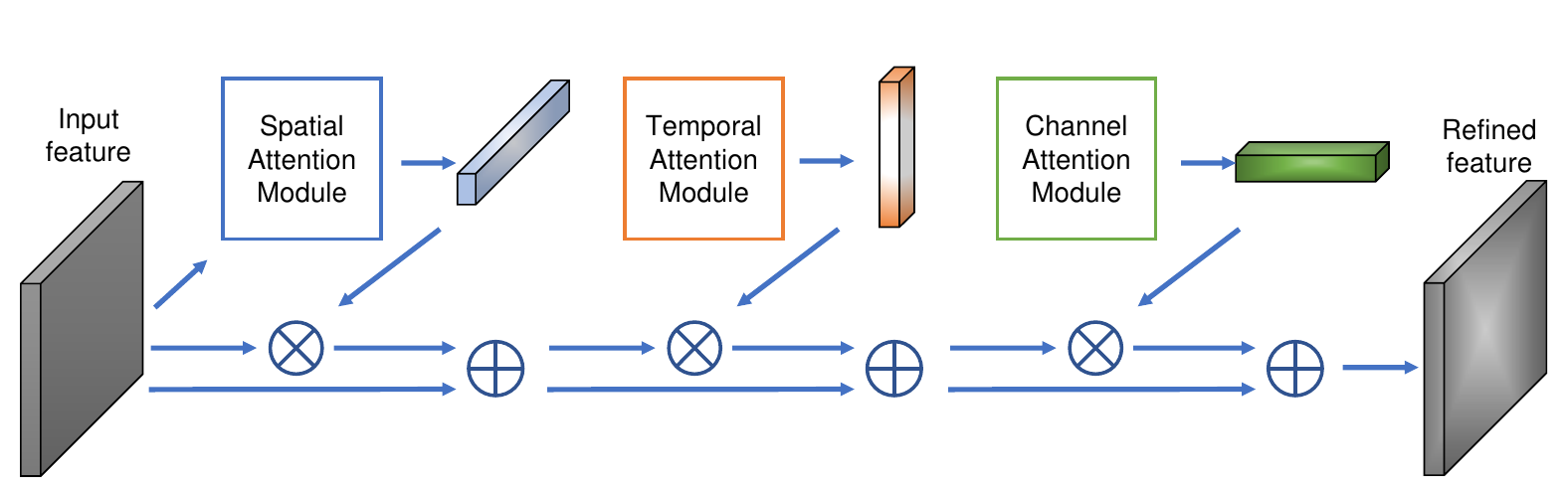}
	\caption{Illustration of the STC-attention module. Three sub-modules are arranged sequentially in the orders of SAM, TAM and CAM. $\otimes$ denotes the element-wise multiplication. $\oplus$ denotes the element-wise addition.}
	\label{fig:stc}	
	
	\end{figure*}

    There have been many formulations for attention modules~\cite{xu_show_2015, wang_residual_2017, vaswani_attention_2017, hu_gather-excite:_2018}. Here, with extensive experiments, we propose a STC-attention module which is shown in Fig.~\ref{fig:stc}. 
    It contains three sub-modules: spatial attention module, temporal attention module and channel attention module. 
    
    \textbf{Spatial attention module (SAM)} can help the model pay different levels of attention for each of the joints. 
    It is computed as:
    \begin{equation}
        \mathbf{M}_s = \sigma(g_s(AvgPool(\mathit{f}_{in}))
        \label{eq:ms}
    \end{equation}
    where $\mathit{f}_{in}\in\mathbb{R}^{C_{in}\times T\times N}$ is the input feature map and is averaged over all of the frames. $g_s$ is a 1-D convolutional operation. $\mathbf{W}_{g_s}\in \mathbb{R}^{1\times C_{in} \times K_s}$ where $K_s$ is the kernel size. $\sigma$ denotes the $Sigmoid$ activation function. 
    The attention map $\mathbf{M}_s\in\mathbb{R}^{1\times 1\times N}$ is then multiplied to the input feature map in a residual manner for adaptive feature refinement.
    
    \textbf{Temporal attention module (TAM)} is similar with the SAM and is computed as:
    \begin{equation}
        \mathbf{M}_t = \sigma(g_t(AvgPool(\mathit{f}_{in}))
        \label{eq:mt}
    \end{equation}
    where $\mathbf{M}_t\in\mathbb{R}^{1\times T\times 1}$ and the definitions of other symbols are similar with  Eq.~\ref{eq:ms}
    
    \textbf{Channel attention module (CAM)} can help models strengthen the discrimitive features (channels) according to the input samples. It generates the attention maps as follows: 
    \begin{equation}
        \label{eq:mc}
        \mathbf{M}_c = \sigma(\mathbf{W}_2(\delta(\mathbf{W}_1(AvgPool(\mathit{f}_{in})))))
    \end{equation}
    where $\mathit{f}_{in}$ is averaged over all of the joints and frames. $\mathbf{M}_c\in\mathbb{R}^{C\times 1\times 1}$. $\mathbf{W}_1\in \mathbb{R}^{(C\times \frac{C}{r})}$ and $\mathbf{W}_2 \in \mathbb{R}^{(\frac{C}{r}\times C)}$ are the weights of two fully-connected layers. $\delta$ denotes the $ReLu$ activation function. 
    
    \textbf{Arrangement of the attention modules:} the three sub-modules introduced above can be placed in different manners: the parallel manner or the sequential manner with different orders. Finally we found that the sequential manner is better, where the order is SAM, TAM and CAM. This will be shown in Sec.~\ref{sec:ablation}
    

    \subsection{Basic block}
\begin{figure}[!htb]
	\centering
	\includegraphics[width=0.7\linewidth]{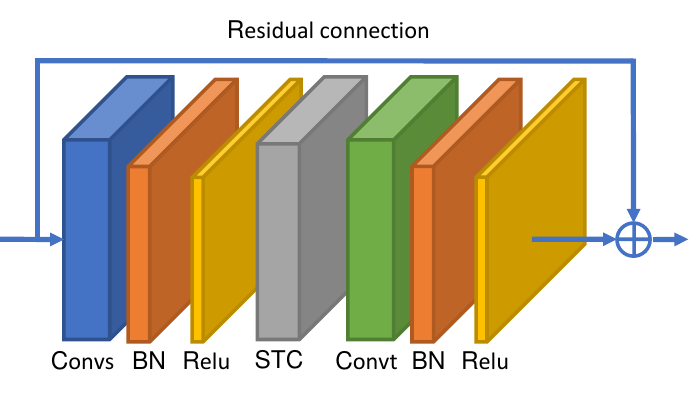}
	\caption{Illustration of the basic block. Convs represents the spatial AGCL, and Convt represents the temporal AGCL, both of which are followed by a BN layer and a ReLU layer. STC represents the STC-attention module. Moreover, a residual connection is added for each block.}
	\label{fig:block}	
	\end{figure}

    The convolution along the temporal dimension is the same as the ST-GCN, i.e., performing the $K_t\times 1$ convolution on the $C\times T \times N$ feature maps. 
    Both the spatial GCN and the temporal GCN are followed by a batch normalization (BN) layer and a ReLU layer. 
    As shown in Fig.~\ref{fig:block}, one basic block is the series of one spatial GCN (Convs),  one STC-attention module (STC) and one temporal GCN (Convt).
    To stabilize the training and ease the gradient propagation, a residual connection is added for each basic block.      
    
	\subsection{Network architecture}
\begin{figure}[!htb]
	\centering
	\includegraphics[width=0.7\linewidth]{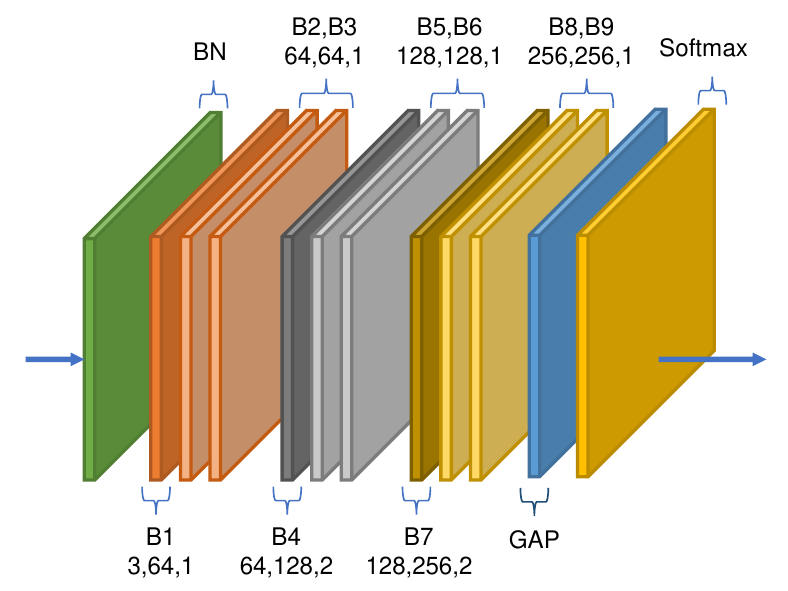}
	\caption{Illustration of the network architecture. There are a total of $9$ basic blocks (B1-B9). The three numbers of each block represent the number of input channels, the number of output channels and the stride, respectively. GAP represents the global average pooling layer.}
	\label{fig:stream}	
	\end{figure}

    As shown in Fig.~\ref{fig:stream}, the overall architecture of the network is the stack of these basic blocks (Fig.~\ref{fig:block}). There are a total of $9$ blocks. 
    The numbers of output channels for each block are $64$, $64$, $64$, $128$, $128$, $128$, $256$, $256$ and $256$. 
    A data BN layer is added at the beginning to normalize the input data.
    A global average pooling layer is performed at the end to pool feature maps of different samples to the same size. 
    The final output is sent to a $softmax$ classifier to obtain the prediction. 
    
    \subsection{Multi-stream networks}
    \label{sec:multistream}
    
    \begin{figure*}[!htb]
	\centering
	\includegraphics[width=0.7\linewidth]{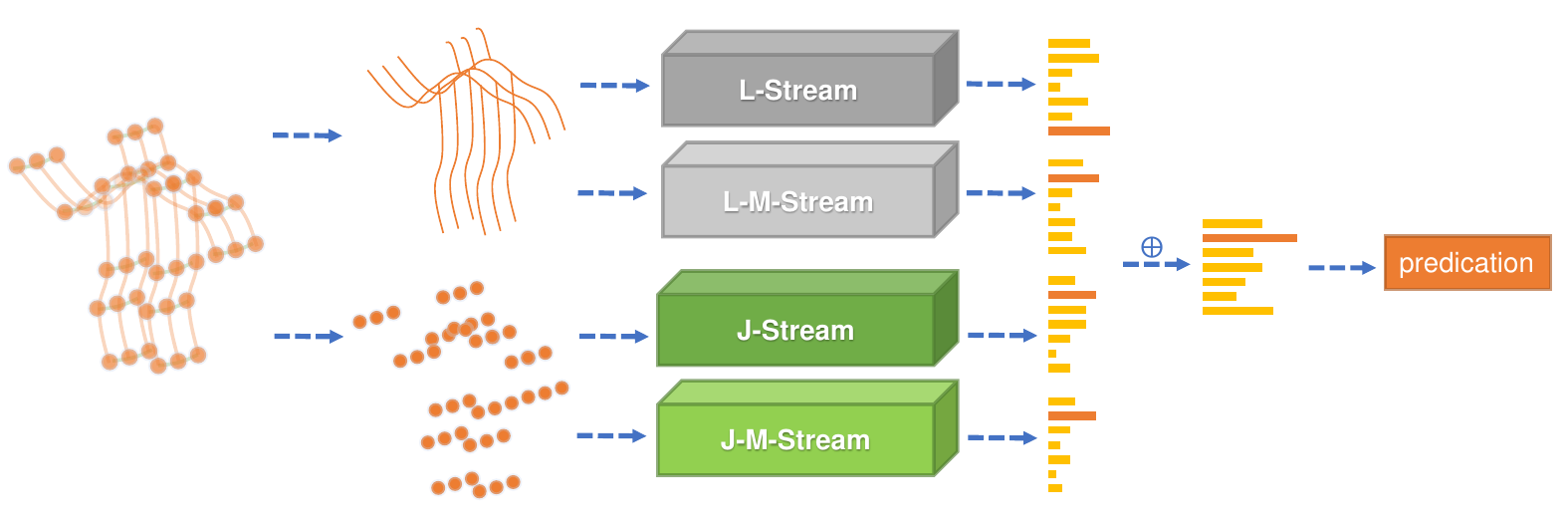}
	\caption{Illustration of the overall architecture of the MS-AAGCN. The $softmax$ scores of the four streams are fused using weighted summation to obtain the final prediction. J denotes the joint information. B denotes the bone information. M denotes the motion information.}
	\label{fig:twostream}	
	\end{figure*}

        
    
    As introduced in Sec.~\ref{sec:introduction}, both the first-order information (the coordinates of the joints) and the second-order information (the direction and length of the bones), as well as their motion information, are worth to be investigated for the skeleton-based action recognition task. 
    In this work, we model these four modalities in a multi-stream framework. 
    
    In particular, we define that the joint closer to the center of gravity of the skeleton is the source joint and the joint farther away from the center of gravity is the target joint. 
    Each bone is represented as a vector pointing from its source joint to its target joint.
    For example, given a bone in frame $t$ with its source joint $\mathbf{v}_{i,t}=(x_{i,t},\ y_{i,t},\ z_{i,t})$ and its target joint $\mathbf{v}_{j, t}=(x_{j, t},\ y_{j, t},\ z_{j, t})$, the vector of the bone is calculated as $\mathbf{e}_{i,j,t} = (x_{j, t}-x_{i,t},\ y_{j, t}-y_{i,t},\ z_{j, t}-z_{i,t})$.
    Since there are no cycles in the graph of the skeleton data, each bone can be assigned with a unique target joint.
    Note that the number of joints is one more than the number of bones because the root joint is not assigned to any bones. 
    To simplify the design of the network, we assign an empty bone with its values as $0$ to the root joint. 
    Thus both the graph and the network of bones can be designed the same as that of joints. 
    We use J-stream and B-stream to represent the networks of joints and bones, respectively. 
    
    As for the motion information, it is calculated as the difference between the same joints or bones in two consecutive frames. 
    For example, given a joint in frame $t$, i.e., $\mathbf{v}_{i,t}=(x_{i,t},\ y_{i,t},\ z_{i,t})$ and the same joint in frame $t+1$, i.e., $\mathbf{v}_{i, t+1}=(x_{i, t+1},\ y_{i, t+1},\ z_{i, t+1})$, the motion information between the $\mathbf{v}_{i,t}$ and $\mathbf{v}_{i, t+1}$ is represented as $\mathbf{m_{i,t,t+1}} = (x_{i, t+1}-x_{i,t},\ y_{i, t+1}-y_{i,t},\ z_{i, t+1}-z_{i,t})$. 

    The overall architecture (MS-AAGCN) is shown in Fig.~\ref{fig:twostream}. 
    The four modalities (joints, bones and their motions) are fed into four streams. 
    Finally, the $softmax$ scores of the four streams are fused using weighted summation to obtain the action scores and predict the action label. 

	\section{Experiments}
	\label{sec:experiments}
\begin{figure}[!htb]
	\centering
	\includegraphics[width=0.8\linewidth]{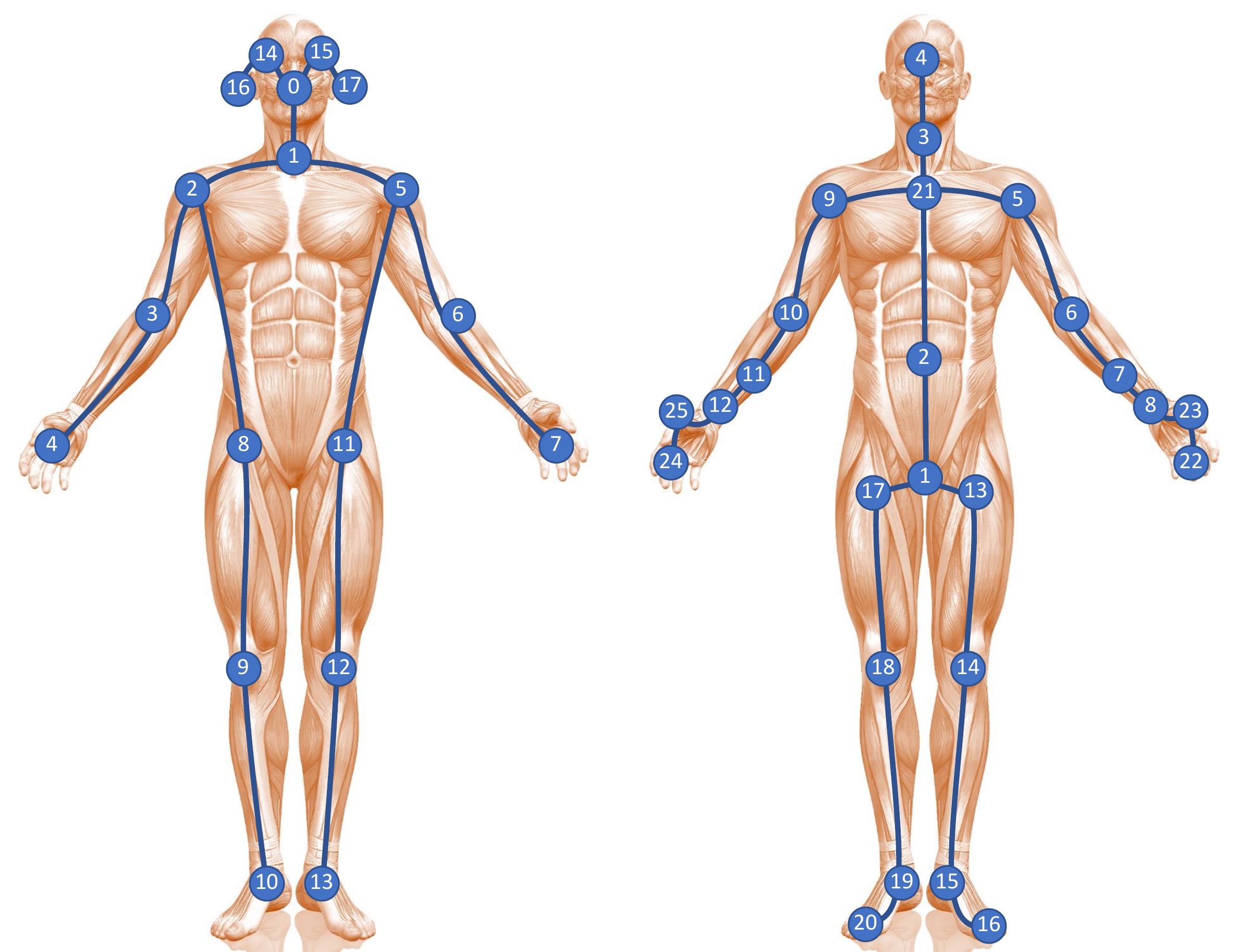}
	\caption{The left sub-graph shows the joint labels of the Kinetics-Skeleton dataset and the right sub-graph shows the joint labels of the NTU-RGBD dataset.}
	\label{fig:dataset}	
	\end{figure}

    To perform a head-to-head comparison with ST-GCN, our experiments are conducted on the same two large-scale action recognition datasets: NTU-RGBD~\cite{shahroudy_ntu_2016} and Kinetics-Skeleton~\cite{kay_kinetics_2017,yan_spatial_2018}. 
    First, since the  NTU-RGBD dataset is smaller than the Kinetics-Skeleton dataset, we perform exhaustive ablation studies on it to verify the effectiveness of the proposed model components based on the recognition performance. 
    Then, the final model is evaluated on both of the datasets to verify the generality and is compared with the other state-of-the-art approaches for skeleton-based action recognition task. 
    Finally, we fuse the skeleton data with pose-guided cropped RGB data and achieve higher accuracies in NTU-RGBD dataset. 
    The definitions of joints and their natural connections in the two datasets are shown in Fig.~\ref{fig:dataset}.
    
    \subsection{Datasets}
    \textbf{NTU-RGBD:} NTU-RGBD~\cite{shahroudy_ntu_2016} is currently the largest and most widely used in-door-captured action recognition dataset, which contains 56,000 action clips in 60 action classes. The clips are performed by 40 volunteers in different age groups ranging from 10 to 35. Each action is captured by 3 cameras at the same height but from different horizontal angles: $-45^\circ, 0^\circ, 45^\circ$. 
    This dataset provides 3D joint locations of each frame detected by Kinect depth sensors. There are 25 joints for each subject in the skeleton sequences, while each video has no more than 2 subjects. The original paper~\cite{shahroudy_ntu_2016} of the dataset recommends two benchmarks: 
    (1) Cross-subject (CS): the dataset in this benchmark is divided into a training set (40,320 videos) and a validation set (16,560 videos), where the subjects in the two subsets are different.
    (2) Cross-view (CV): the training set in this benchmark contains 37,920 videos that are captured by cameras 2 and 3, and the validation set contains 18,960 videos that are captured by camera 1.
    We follow this convention and report the top-1 accuracy on both benchmarks.
    
    \textbf{Kinetics-Skeleton:} Kinetics~\cite{kay_kinetics_2017} is a large-scale human action dataset that contains 300,000 videos clips in 400 classes. The video clips are sourced from YouTube videos and have a great variety. It only provides raw video clips without skeleton data. 
   ~\cite{yan_spatial_2018} estimate the locations of 18 joints on every frame of the clips using the publicly available OpenPose toolbox~\cite{cao_realtime_2017}. Two peoples are selected for multi-person clips based on the average joint confidence. 
    We use their released data (Kinetics-Skeleton) to evaluate our model. 
    The dataset is divided into a training set (240,000 clips) and a validation set (20,000 clips). Following the evaluation method in~\cite{yan_spatial_2018}, we train the models on the training set and report the top-1 and top-5 accuracies on the validation set.
	
    \subsection{Training details}
    \label{sec:trainingdetails}
     All experiments are conducted on the PyTorch deep learning framework~\cite{paszke_automatic_2017}.
     Stochastic gradient descent (SGD) with Nesterov momentum ($0.9$) is applied as the optimization strategy. The batch size is $64$. Cross-entropy is selected as the loss function to back-propagate gradients. The weight decay is set to $0.0001$.
     
    For the NTU-RGBD dataset, there are at most two peoples in each sample of the dataset. If the number of bodies in the sample is less than 2, we pad the second body with 0. The max number of frames in each sample is 300. 
    For samples with less than 300 frames, we repeat the samples until it reaches 300 frames. The learning rate is set as $0.1$ and is divided by $10$ at the $30_{th}$ epoch and $40_{th}$ epoch. The training process is ended at the $50_{th}$ epoch. 
     
    For the Kinetics-Skeleton dataset, the size of the input tensor of Kinetics is set the same as~\cite{yan_spatial_2018}, which contains 150 frames with 2 bodies in each frame. 
    We perform the same data-augmentation skills as in~\cite{yan_spatial_2018}.
    In detail, we randomly choose 150 frames from the input skeleton sequence and slightly disturb the joint coordinates with randomly chosen rotations and translations. The learning rate is also set as $0.1$ and is divided by $10$ at the $45_{th}$ epoch and $55_{th}$ epoch. The training process is ended at the $65_{th}$ epoch. 
     
    \subsection{Ablation Study}
    \label{sec:ablation}
    We verify the effectiveness of the proposed components in MS-AAGCN in this section using the NTU-RGBD dataset. 
    
    \subsubsection{Learning rate scheduler and data preprocessing}
	In the original paper of ST-GCN, the learning rate is multiplied by $0.1$ at the $10_{th}$ and $50_{th}$ epochs. The training process is ended in $80$ epochs. We rearrange the learning rate scheduler from [10, 50, 80] to [30, 40, 50] and obtain the better performance (shown in Tab.~\ref{tab:preprocess}, ``before preprocessing"). 

\begin{figure}[!htb]
    \centering
	\includegraphics[width=1\linewidth]{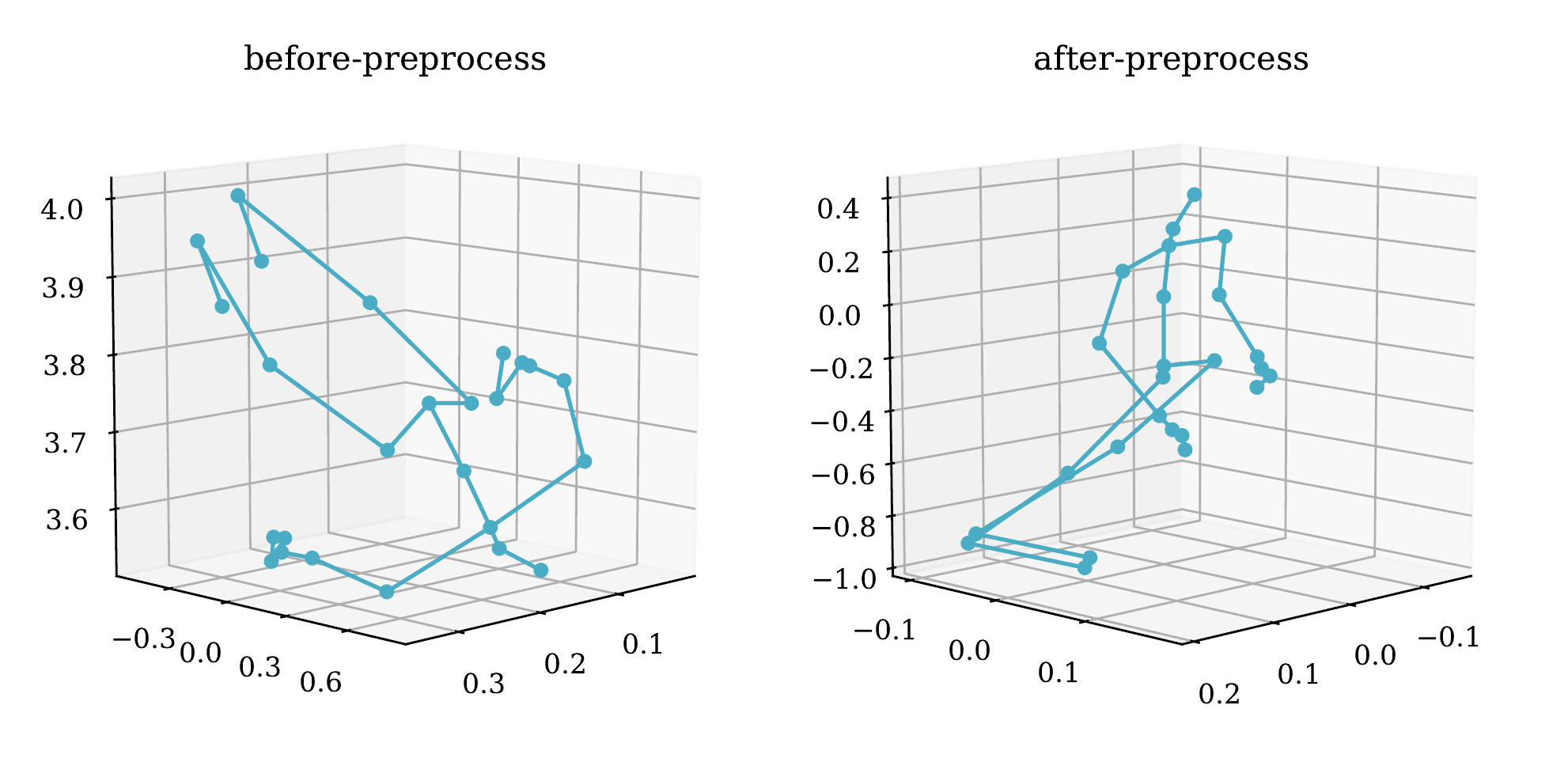}
	\caption{Example of the data preprocessing on the NTU-RGBD dataset. The left is the original skeleton, and the right is the preprocessed skeleton.}
	\label{fig:rot}    
	\end{figure}

    Moreover, we use some preprocessing strategies on the NTU-RGBD dataset. The body tracker of Kinect is prone to detecting more than 2 bodies, some of which are objects. To filter the incorrect bodies, we first select two bodies in each sample based on the body energy. The energy is defined as the average of the skeleton's standard deviation across each of the channels. 
    Subsequently, each sample is normalized to make the distribution of the data for each channel unified. In detail, the coordinates of each joint are subtracted from the coordinates of the ``spine joint" (the $2_{nd}$ joint of the left sub-graph in Fig.~\ref{fig:dataset}). 
    Finally, as different samples may be captured in different viewpoints,  similar to~\cite{shahroudy_ntu_2016}, we translate the original 3D location of the body joints from the camera coordinate system to body coordinates. 
    For each sample, we perform a 3D rotation to fix the $X$ axis parallel to the 3D vector from the ``right shoulder" ($5_{th}$ joint) to the ``left shoulder" ($9_{th}$ joint), and the $Y$ axis toward the 3D vector from the ``spine base” ($21_{st}$ joint) to the “spine” ($2_{nd}$ joint). Fig.~\ref{fig:rot} shows an example of the preprocessing. 
    The performance after the preprocessing is referred as ``after-preprocessing" in Tab.~\ref{tab:preprocess}. 
    
	Tab.~\ref{tab:preprocess} shows that the preprocessing considerably helps the recognition. It is because that the original data are too noisy.
    
	\begin{table}[htb]
      \centering
		\begin{tabular}{lcc}
			\hline
			Methods     & CS ($\%$)  & CV ($\%$)   \\
			\hline
			original performance in~\cite{yan_spatial_2018} & 81.5& 88.3        \\
            before preprocessing & 82.4 & 90.1        \\
			after preprocessing  & 84.3   & 92.7          \\
			\hline
		\end{tabular}
    \caption{Comparisons of the validation accuracy using rearranged learning-rate scheduler and data preprocessing.}
    \label{tab:preprocess}
	\end{table}
    
    \subsubsection{Adaptive graph convolutional block}

    We use the ST-GCN as the baseline method.
    As introduced in Section~\ref{sec:agcn}, there are $2$ kinds of sub-graphs in the our proposed AGCL, i.e., the global graph $B$ and the individual graph $C$. 
    We test the performance of using each of the graphs along and combining them together. The results are shown as AGCN-B, AGCN-C and AGCN-BC  in Tab.~\ref{tab:nlgcn}, respectively. 
    It shows that both of the two designed graphs brings notable improvement for the action recognition task. With two graphs added together, the model obtains the best performance. 
    
    Besides, the performance of initial strategies introduced in Sec.~\ref{sec:init} is tested and shown as  AGCN-ABC and AGCN-BC in Tab.~\ref{tab:nlgcn}. It suggests the second strategy is slightly better. 
    
    Moreover, we verify the effectiveness of adding the gating mechanism (AGCN-BC-G), which also brings encouraging improvement. Overall, the complete AGCL brings improvements of $+3.1\%$ and $+1.7\%$ on CS and CV benchmarks, respectively. 
    
	\begin{table}[htb]
 	\centering
		\begin{tabular}{lcc}
			\hline
			Methods     & CS ($\%$) & CV ($\%$)\\
			\hline
            STGCN    & 84.3  & 92.7      \\   
            AGCN-B   & 86.4  & 93.6       \\
			AGCN-C   & 86.1  & 93.5      \\ 
			AGCN-ABC & 86.6  & 93.7      \\ 
			AGCN-BC  & 87.0  & 94.1      \\ 
			AGCN-BC-G  & 87.4  & 94.4      \\ 
			\hline
		\end{tabular}
     	\caption{Comparisons of the validation accuracies on the NTU-RGBD dataset. 
     	$A$ denotes the adjacency matrix of the body-based graph shown in Eq.~\ref{eq:stgcni}.
     	$B$ and $C$ denote the global graph and the individual graph introduced in Sec~\ref{sec:agcn}, respectively. $G$ denotes using the gating mechanism.}
        \label{tab:nlgcn}
	\end{table}

    \subsubsection{Attention module}
    In this section, we verify the effectiveness of the proposed STC-attention module introduced in Sec.~\ref{sec:attention}. 
    The results are shown in Tab.~\ref{tab:attention}. 
    We first separately test the contributions of three sub-modules (SAM, TAM and CAM) based on the ST-GCN , shown as ASTGCN-S, ASTGCN-T and ASTGCN-C, respectively. 
    It shows that all of the three sub-modules can help improving the performance.
    Then we test the performance of adding each of the sub-modules as well as concatenating them sequentially, shown as STGCN-ADD and STGCN-STC, respectively. 
    It suggests that concatenating the three sub-modules is slightly better. 
    Finally, we embed the STC-attention module into the AGCN and obtains the similar results. 
    Note that the improvement brought by the attention module for AGCN ($+0.6\%$ and $+0.7\%$) is less significant than that for STGCN ($+2.8\%$ and $+1.5\%$). 
    We argue that it is because the AGCN is powerful and the accuracy is already very high, thus the effect of the attention module is limited. 

    \begin{table}[htb]
 	\centering
		\begin{tabular}{lcc}
			\hline
			Methods     & CS ($\%$) & CV ($\%$)\\
			\hline
			STGCN        & 84.3  & 92.7      \\ 
			ASTGCN-S      & 86.1  & 93.1      \\ 
			ASTGCN-T      & 86.6  & 93.6      \\ 
			ASTGCN-C      & 86.5  & 93.7      \\ 
			ASTGCN-ADD    & 86.8  & 94.1      \\
			ASTGCN-STC    & 87.1  & 94.2      \\
			\hline
			AGCN       & 87.4  & 94.4      \\ 
			AAGCN-ADD  & 87.7  & 94.8      \\ 
			AAGCN-STC  & 88.0  & 95.1      \\ 
			\hline
		\end{tabular}
     	\caption{Comparisons of the validation accuracy on the NTU-RGBD dataset for each of the attention sub-modules and different arrangement strategies. STGCN and AGCN denote the STGCN-M and AGCN-BC-G in Tab.~\ref{tab:nlgcn}, respectively. STC means concatenating the three sub-modules sequentially.}
        \label{tab:attention}
	\end{table}

    \subsubsection{Multi-stream framework}
    Finally we test the performance of using four proposed data modalities and show the results in Tab.~\ref{tab:mulitistream}. 
    Here, J, B, J-M and B-M denote the joint modality, the bone modality, the motion of joint and the motion of bone introduced in Sec.~\ref{sec:multistream}, respectively.
    Clearly, the multi-stream method outperforms the single-stream methods.
    For single-stream methods, the bone modality (B-AAGCN) performs slightly better than the joint modality (J-AAGCN) for CS benchmark. 
    As for the CV benchmark, the result is reversed. 
    This suggests the complementarity of the two modalities. By combing the joints and bones (JB-AAGCN), it brings notable improvement as expected. 
    
    The performance of motion modalities (J-M-AAGCN and B-M-AAGCN) is generally lower than the performance of the joint and bone modalities. 
    However, adding them together still brings improvement.
    
    \begin{table}[htb]
		\centering
		\begin{tabular}{lcc}
			\hline
			Methods     & CS ($\%$) & CV ($\%$)\\
			\hline
			J-AAGCN & 88.0 & 95.1      \\
			B-AAGCN & 88.4 & 94.7       \\
			J-M-AAGCN & 85.9 & 93.0      \\
			B-M-AAGCN & 86.0 & 93.1       \\
            \hline
            JB-AAGCN & 89.4 & 96.0        \\
            MS-AAGCN & 90.0 & 96.2        \\
			\hline
		\end{tabular}
        \caption{Comparisons of the validation accuracy with different input modalities on the NTU-RGBD dataset. JB represents using the joint and bone modalities. MS represents using all of the four modalities. AAGCN denotes the AAGCN-STC in Tab.~\ref{tab:attention}.}
		\label{tab:mulitistream}
	\end{table}

    \subsection{Comparisons with the state-of-the-art methods}
    We compare the final model with the state-of-the-art skeleton-based action recognition methods on both the NTU-RGBD dataset and Kinetics-Skeleton dataset. 
    The results are shown in Tab.~\ref{tab:ntu-rgbd} and Tab.~\ref{tag:kinetics}, respectively. 
    The methods used for comparisons include the handcraft-feature-based methods~\cite{vemulapalli_human_2014,fernando_modeling_2015}, RNN-based methods~\cite{du_hierarchical_2015,shahroudy_ntu_2016,liu_spatio-temporal_2016,song_end--end_2017,zhang_view_2017,li_skeleton-based_2018,li_independently_2018, si_skeleton-based_2018}, CNN-based methods~\cite{kim_interpretable_2017,ke_new_2017,liu_enhanced_2017,li_skeleton-based_2017,li_skeleton_2017} and GCN-based methods~\cite{yan_spatial_2018,tang_deep_2018}.
    Our model achieves the state-of-the-art performance with a large margin on both of the datasets, which suggests the superiority of our model.
    
      \begin{table}[htb]
      \centering
		\begin{tabular}{lcc}
			\hline
			Methods     &CS (\%)& CV (\%)    \\
            \hline
            Lie Group~\cite{vemulapalli_human_2014} & 50.1 & 82.8 \\
			\hline
            HBRNN~\cite{du_hierarchical_2015} & 59.1  &    64.0  \\
            Deep LSTM~\cite{shahroudy_ntu_2016}   & 60.7 &    67.3  \\
            ST-LSTM~\cite{liu_spatio-temporal_2016} & 69.2 & 77.7 \\
            STA-LSTM~\cite{song_end--end_2017}  & 73.4 &    81.2  \\
            VA-LSTM~\cite{zhang_view_2017}  & 79.2 &    87.7  \\
            ARRN-LSTM~\cite{li_skeleton-based_2018} & 80.7 & 88.8 \\
            Ind-RNN~\cite{li_independently_2018} & 81.8 & 88.0 \\
            SRN+TSL~\cite{si_skeleton-based_2018} & 84.8 & 92.4 \\
            \hline
            TCN~\cite{kim_interpretable_2017}   & 74.3&    83.1      \\
            Clips+CNN+MTLN~\cite{ke_new_2017} & 79.6 & 84.8 \\
            Synthesized CNN~\cite{liu_enhanced_2017}   & 80.0 &      87.2  \\
            CNN+Motion+Trans~\cite{li_skeleton-based_2017} & 83.2  &      89.3  \\
            3scale ResNet152~\cite{li_skeleton_2017}  & 85.0  &      92.3  \\
            \hline
            ST-GCN~\cite{yan_spatial_2018} & 81.5  &      88.3  \\
            DPRL+GCNN~\cite{tang_deep_2018} & 83.5 & 89.8 \\
            \hline
            MS-AAGCN (ours)&  90.0 & 96.2 \\
			\hline
		\end{tabular}
      \caption{Comparisons of the validation accuracy with state-of-the-art methods on the NTU-RGBD dataset.}
      \label{tab:ntu-rgbd}
	\end{table}
    
    \begin{table}[htb]
    \centering
		\begin{tabular}{lcc}
			\hline
			Methods     & Top-1 (\%)& Top-5 (\%)    \\
			\hline
			Feature Enc.~\cite{fernando_modeling_2015} & 14.9  &    25.8  \\
			Deep LSTM ~\cite{shahroudy_ntu_2016}   & 16.4 &    35.3  \\
			TCN ~\cite{kim_interpretable_2017}   & 20.3&    40.0      \\
            ST-GCN ~\cite{yan_spatial_2018}  & 30.7 &      52.8  \\
            \hline
            J-AAGCN (ours)& 36.0& 58.4\\
            B-AAGCN (ours)& 34.7& 57.5\\
            J-M-AAGCN (ours)& 31.7& 54.6\\
            B-M-AAGCN (ours)& 29.7& 50.0\\
            JB-AAGCN (ours)&  37.4& 60.4\\
            MS-AAGCN (ours)&  37.8& 61.0\\
			\hline
		\end{tabular}
	\caption{Comparisons of the validation accuracy with state-of-the-art methods on the Kinetics-Skeleton dataset.}
    \label{tag:kinetics}
	\end{table}

    \section{Fusion with the RGB modality}
    \label{sec:fusion}
    The skeleton data is robust to the dynamic circumstance and complicated background. 
    However, it lacks the appearance information. 
    For example, if a person is eating something, it is hard to judge whether he is eating an apple or a peal using only the skeleton data. 
    
    In this section, we investigate the necessity of fusing both the skeleton data and the RGB data for the action recognition task on the NTU-RGBD dataset.
    We use a two-stream framework where one of the stream models the RGB data with the 3D convolutional networks and another stream models the skeleton data with our MS-AAGCN. 
    The details of skeleton-stream is the same as Sec~\ref{sec:trainingdetails}.
    For RGB-stream, inspired by~\cite{shi_gesture_2019}, we use the ResNeXt3D-101 model that is pre-trained on ImageNet and Kinetics. 
    When training, we randomly crop a clip from the whole video whose length is random sampled from $[32, 64, 128]$. The crop position is randomly selected from four corners and one center. Then each image is cropped based on the crop ratio sampled form $[1, 0.875, 0.75]$. Note that the width and height of the cropped image can be different. Finally the cropped sequence of images are normalized and resized to $[16, 224, 224]$, which denotes the length, height and width of the clip. The learning rate is initialized with $0.01$ and is multiplied with $0.1$ after the validation accuracy saturates. We use four TITANXP-GPUs and the batch size is set to $32$. Momentum-SGD is used as the optimizer and the weight decay is set to $0.0005$. 
    When testing, we crop clips with different lengths and sizes in different positions. The result is the average of these clips.
    
    Moreover, to get rid of the interference of the background, we propose to crop the persons from the original images and use only the person part for the recognition. 
    In detail, we calculate the border of the persons in each image and use their union as the crop box. 
    This is referred as the pose-guided cropping strategy. 
    
    Tab.~\ref{tab:rgb} shows the results of our methods as well as the previous methods that also use the RGB modality. 
    Here, C denotes using the pose-guided cropping strategy. 
    It shows that the pose-guided cropping strategy brings notable improvement when using only the RGB modality (RNX3D101 vs RNX3D101-C). 
    This suggests that the model using only RGB data is easily misguided by the complicated background. 
    However, the improvement decreases when fusing the skeleton data and the RGB data (RNX3D101+MS-AAGCN vs RNX3D101+MS-AAGCN-C). 
    The reason is that the skeleton data can effectively avoid the interference of the circumstance, thus the contribution of the cropping strategy becomes not as obvious as before. 
    Our best model, i.e., RNX3D101+MS-AAGCN-C, achieves $96.1\%$ for CS benchmark and $99.0\%$ for CV benchmark.
    As shown in Tab.~\ref{tab:rgb}, it exceeds the previous methods with a large margin. 

    \begin{table}[htb]
    \centering
		\begin{tabular}{lcccc}
			\hline
			Methods    & Pose & RGB &CS (\%)& CV (\%)    \\
			\hline
			DSSCA-SSLM~\cite{shahroudy_deep_2018}  &\checkmark &\checkmark & 74.9 &    -  \\
			Chained Network ~\cite{zolfaghari_chained_2017}   &\checkmark &\checkmark& 80.8 &    -  \\
			RGB + 2D Pose~\cite{luvizon_2d/3d_2018} &\checkmark &\checkmark& 85.5  &    -  \\
			Glimpse Clouds ~\cite{baradel_glimpse_2018}   & &\checkmark& 86.6&    93.2      \\
            PEM ~\cite{liu_recognizing_2018}  &\checkmark &\checkmark& 91.2 &      95.3  \\			I3D+RNN+Attention ~\cite{das_where_2019}  &\checkmark &\checkmark & 93.0&    95.4      \\
            \hline
            MS-AAGCN            &\checkmark & & 90.0 & 96.2 \\
            RNX3D101            & &\checkmark & 85.3 & 92.6 \\
            RNX3D101-C        & &\checkmark & 94.6 & 97.0 \\
            \hline
            RNX3D101+MS-AAGCN     &\checkmark &\checkmark&  95.5 & 98.0 \\
            RNX3D101+MS-AAGCN-C &\checkmark &\checkmark&  96.1 & 99.0 \\
			\hline
		\end{tabular}
	\caption{Comparisons of the validation accuracy using RGB modality on the NTU-RGBD dataset.}
    \label{tab:rgb}
	\end{table}

    \section{Visualization and discussion}
    \label{sec:visualization}
    \subsection{Adaptive graphs}
\begin{figure}[!htb]
	\centering
	\includegraphics[width=0.95\linewidth]{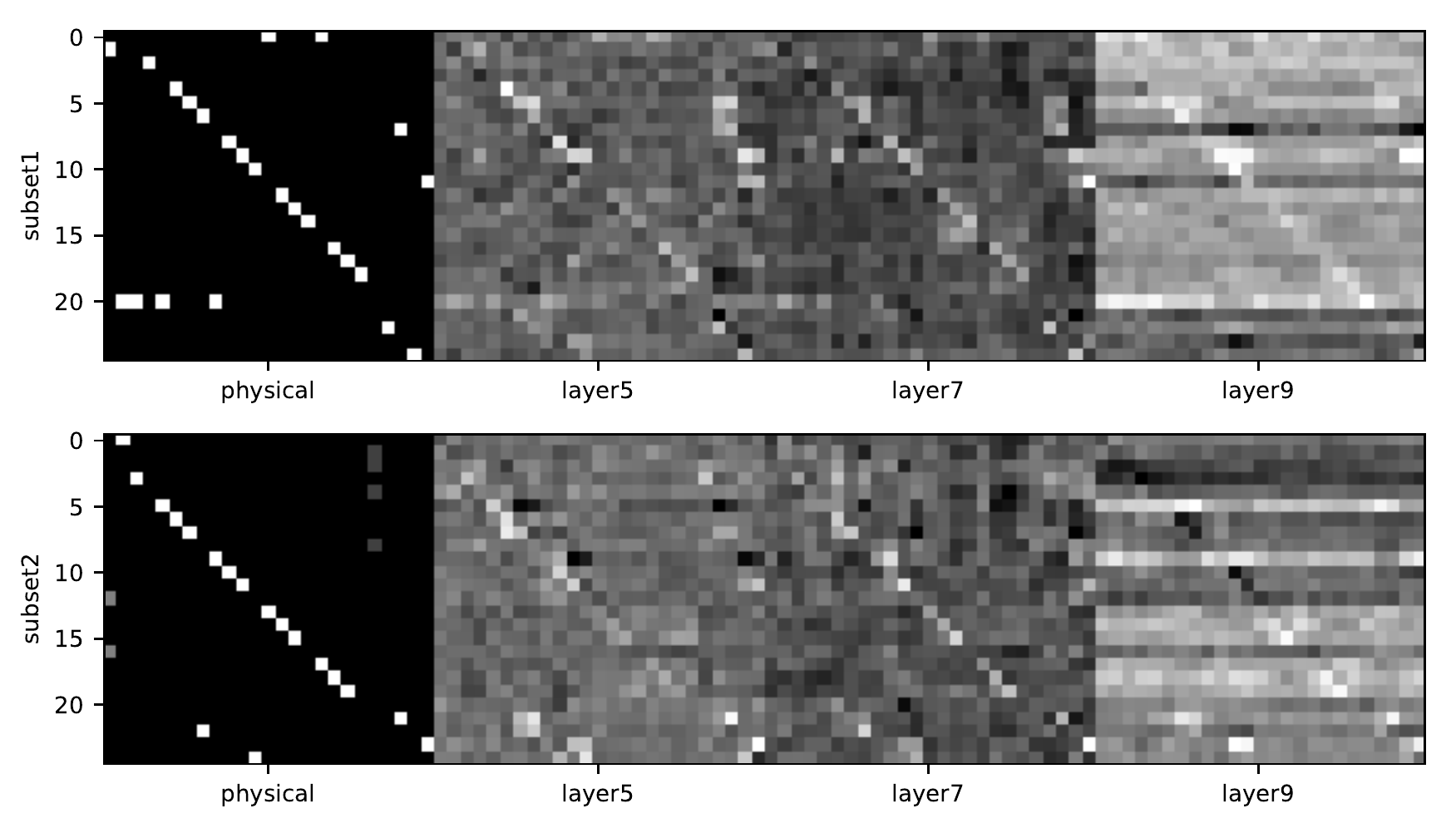}
	\caption{Example of the learned adjacency matrices of the global graph. Different rows shows different subsets. The first column is the adjacency matrices of the human-body-based graph in the NTU-RGBD dataset. Others are examples of the learned adaptive adjacency matrices of different layers learned by our model. 
    }
	\label{fig:graphb}	
	\end{figure}

    The are two kinds of graphs in our model: the global graph and the individual graph. 
    Fig.~\ref{fig:graphb} shows an example of the learned adjacency matrices of the global graph for different subsets and different layers.
    The first and second rows show the adjacency matrices of the centripetal subset ($\mathcal{S}_{i2}$) and the centrifugal subset ($\mathcal{S}_{i3}$) introduced in Sec~\ref{sec:graphconvolution}, respectively. 
    The first column shows the graph structure defined based on the natural connectivity of the human body, i.e., $\mathbf{A}$ in Eq.~\ref{eq:stgcni}. Others are the adjacency matrices of the global graph in different layers. 
	The gray scale of each element in the matrix represents the strength of the connection.

	It shows that the topology of the learned graph is updated based on the human-body-based graph but has many changes. 
	It confirms that the human-body-based graph is not the optimal choice for the action recognition task. 
	Besides, it shows that the graph topology of the higher layers changes more than the lower layers. It is because the information contained in higher layers is more semantic, thus the required topology of graph is more different from the human-body-based graph. 
	
	\begin{figure}[!htb]
	\centering
	\includegraphics[width=0.95\linewidth]{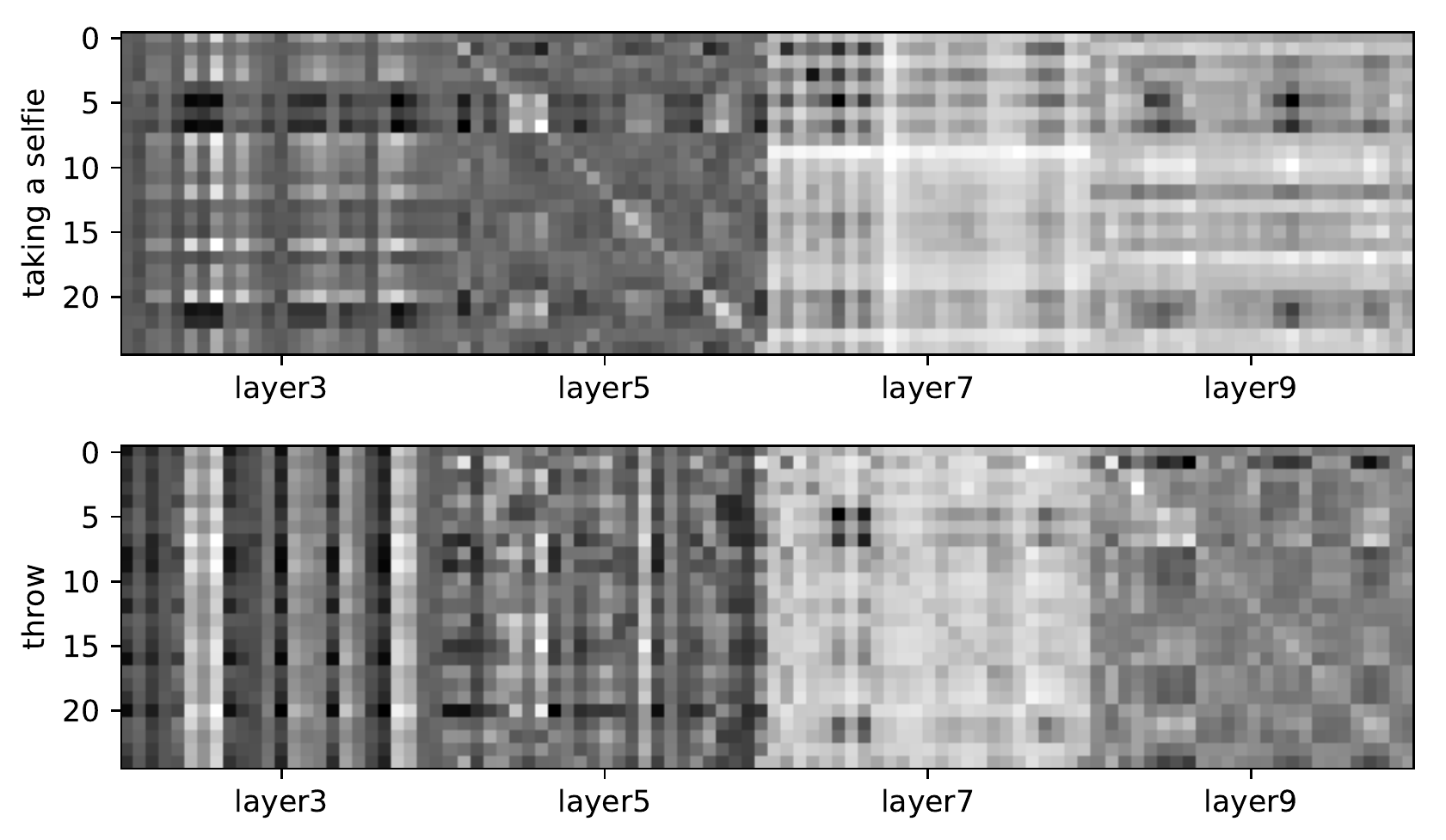}
	\caption{Examples of the learned adjacency matrices of the individual graph. The first and second rows show different samples. Different columns represent different layers.  
    }
	\label{fig:graphc}	
	
	\end{figure}
	
	Similarly, Fig.~\ref{fig:graphc} shows some examples of the learned adjacency matrices of the individual graph for two different samples. 
	It shows that different samples and layers need different graph topologies, which confirms our motivation. 
	
\begin{figure}[!htb]
	\centering
	\includegraphics[width=0.95\linewidth]{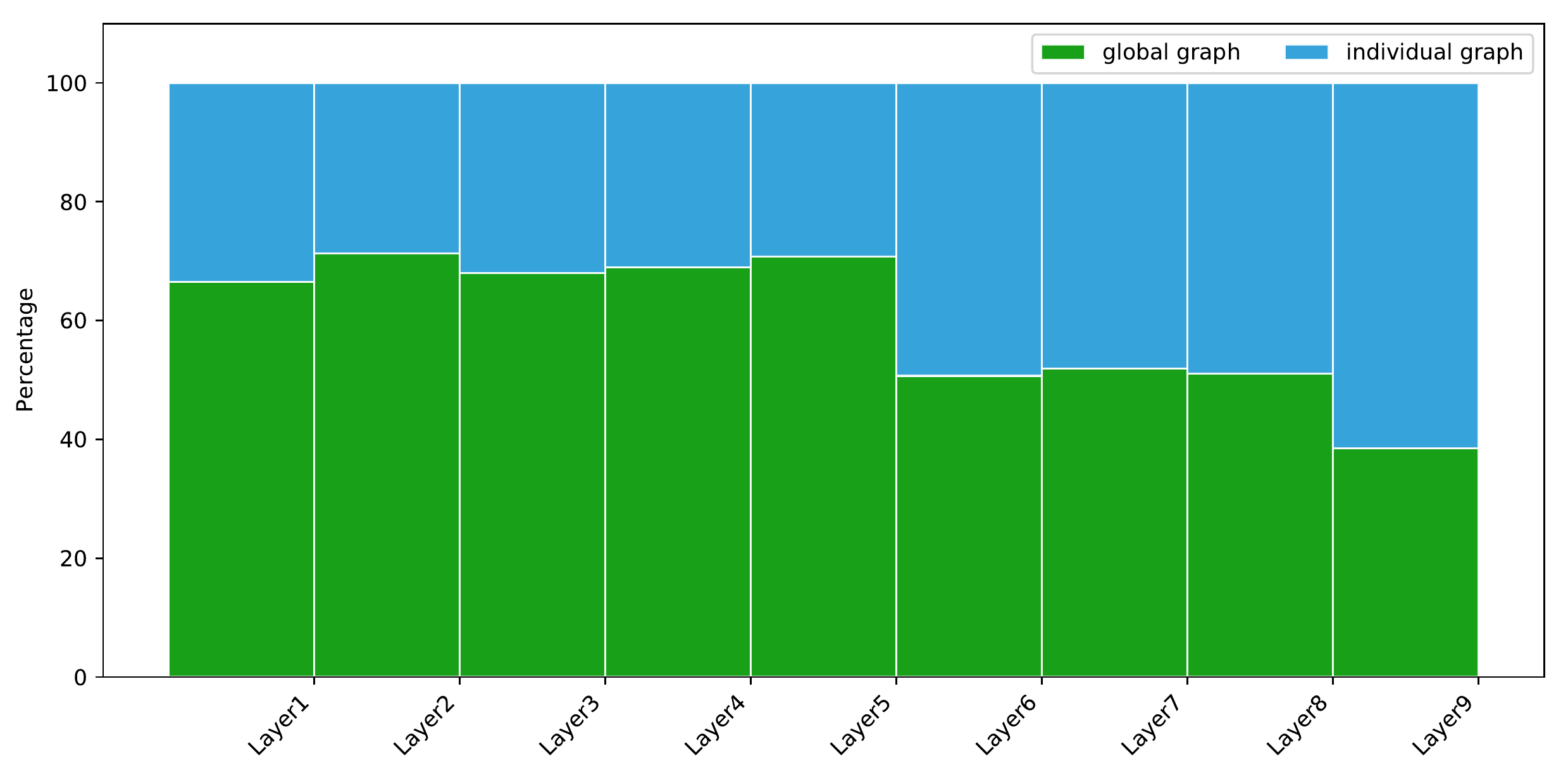}
	\caption{Visualization for the importance of the two kinds of graphs in each of the layers. 
    }
	\label{fig:ratio}	
	
	\end{figure}

	The two kinds of the graphs are fused using the gating mechanism. We visualize the importance of the two kinds of graphs in each layer in Fig.~\ref{fig:ratio}. 
	It shows that the individual graph is more important in top layers.
	This is because the individual graph is learned based on the features of the data samples. The receptive fields of the top layers are larger and their features are more informative. 
	Thus it is easier for top layers to learn the topology of the individual graph than lower layers. 
	
\begin{figure}[!htb]
	\centering
	\includegraphics[width=0.95\linewidth]{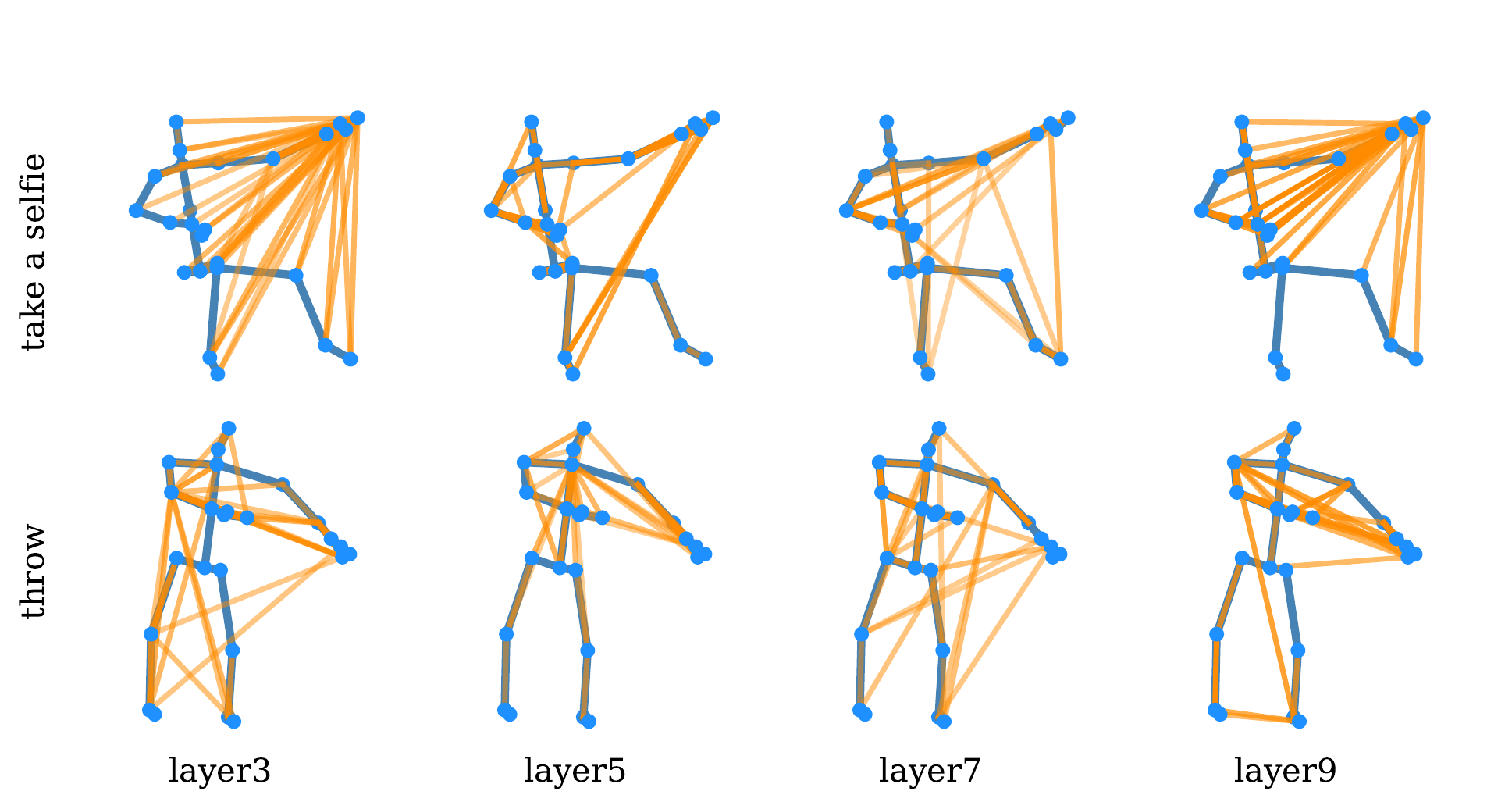}
	\caption{Examples of the learned graph topologies. The orange lines represent the connections whose values are in the top 50.}
	\label{fig:graphbc}	
	
	\end{figure}

	To see the learned graph topologies clearly, we draw the connections between joints on the skeleton sketches as shown in Fig.~\ref{fig:graphbc}. 
	The orange lines represent the connections whose values are in the top 50. 
	The alpha value of the lines represents the strength of the connections. 
	It shows that the learned graph topology is accorded with the human intuition. 
	For example, the relationships between the hand and the head should be more important to recognize the action of ``take a selfie", and the learned graphs did have more connections between them as shown in the first row. 
	This confirms the effectiveness and necessity of the proposed adaptive graph convoutional layer. 

    \subsection{Attention module}
\begin{figure}[!htb]
	\centering
	\includegraphics[width=0.95\linewidth]{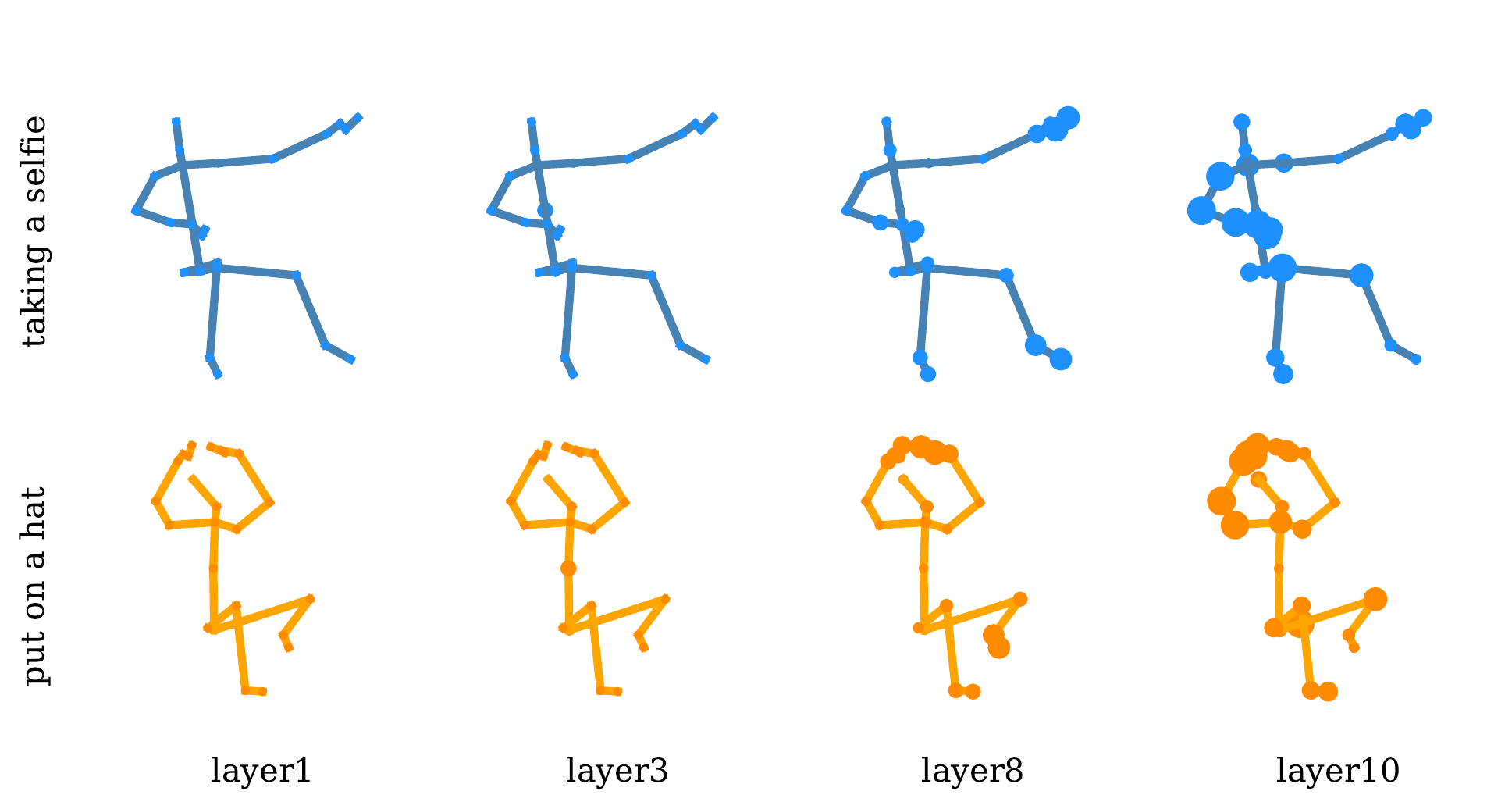}
	\caption{Examples of the learned spatial attention maps. The size of the circle represents the importance of the corresponding joint.
    }
	\label{fig:sa}	
	
	\end{figure}

	There are three sub-modules for our proposed STC-attention module: the spatial attention module, the temporal attention module and the channel attention module. 
	For the spatial attention, we show the learned attention map for different samples and different layers in Fig.~\ref{fig:sa}. 
	The size of the circle represents the importance of the corresponding joint. 
	It shows that the model focus more on the joints of hands and head. Besides, the attention is not obvious in the lower layers. It is because the receptive fields of lower layers are relatively smaller, thus is hard to learn good attention maps.
	
\begin{figure}[!htb]
	\centering
	\includegraphics[width=0.99\linewidth]{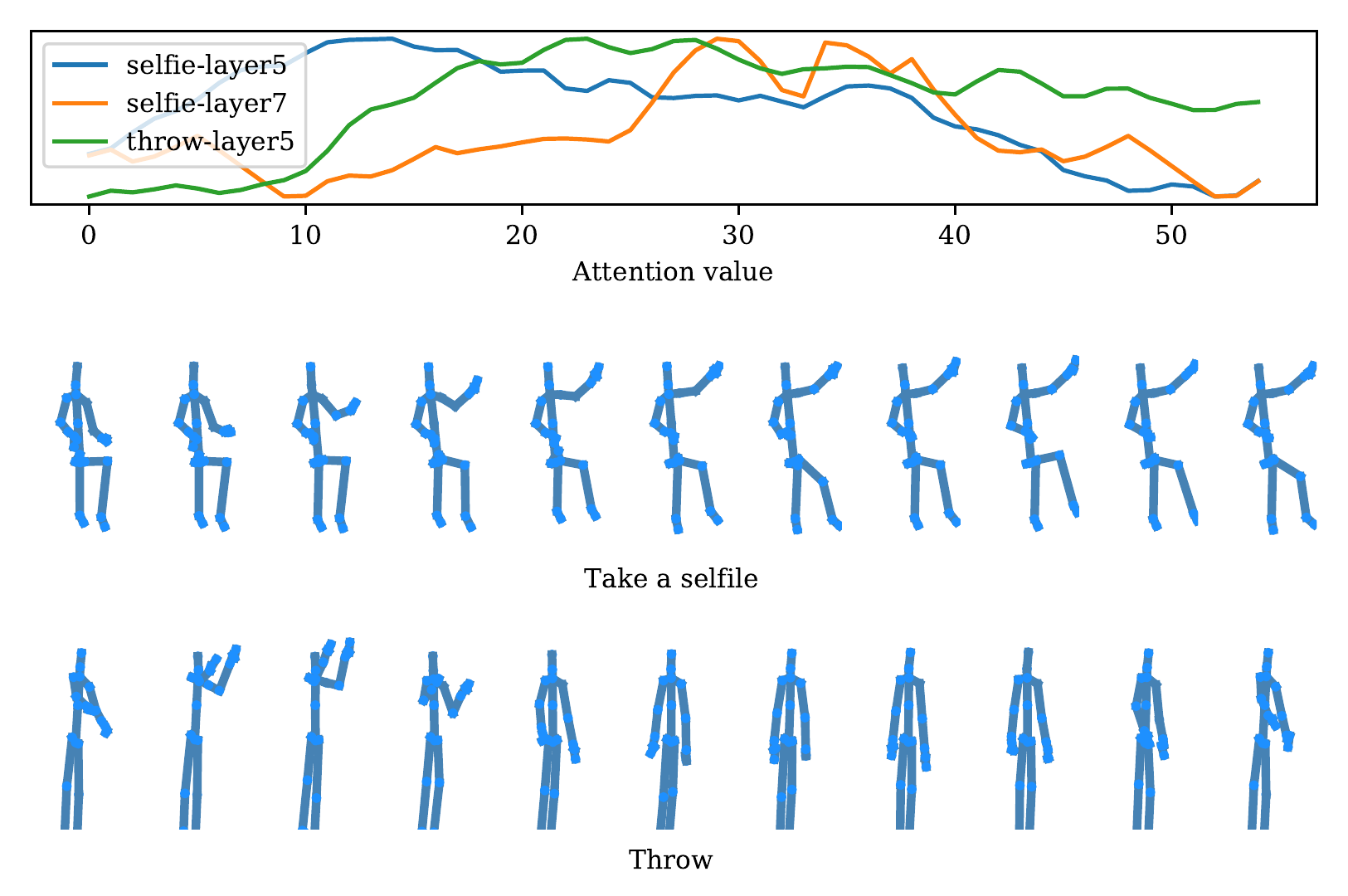}
	\caption{Visualization of the temporal attention map. The first row shows the learned temporal attention weights for each of the frames for different layers and samples. The second and third rows show the corresponding skeleton sketches.}
	\label{fig:ta}	
	
	\end{figure}

	For the temporal attention, we show an example of the learned attention weights for each of the frames and the corresponding skeleton sketches in Fig.~\ref{fig:ta}. 
	For the sample of taking a selfie, it shows the model focus more on the process of raising people's hand in the fifth layer, and cares more about the final posture of the selfie in the seventh layer. 
	For the sample of throwing, although in the same fifth layer, it learns different structure and pays more attentions on the frames where the hands are in the lower position. 
	This shows the effectiveness and the adaptability of the designed module.

	\subsection{Multi-modalities}
    To show the complementarity between different modalities, we plot the accuracy difference for some of the classes between the different modalities. 
    We use the CV benchmark of the NTU-RGBD dataset.
 	
	\begin{figure}[!htb]
	\centering
	\includegraphics[width=0.9\linewidth]{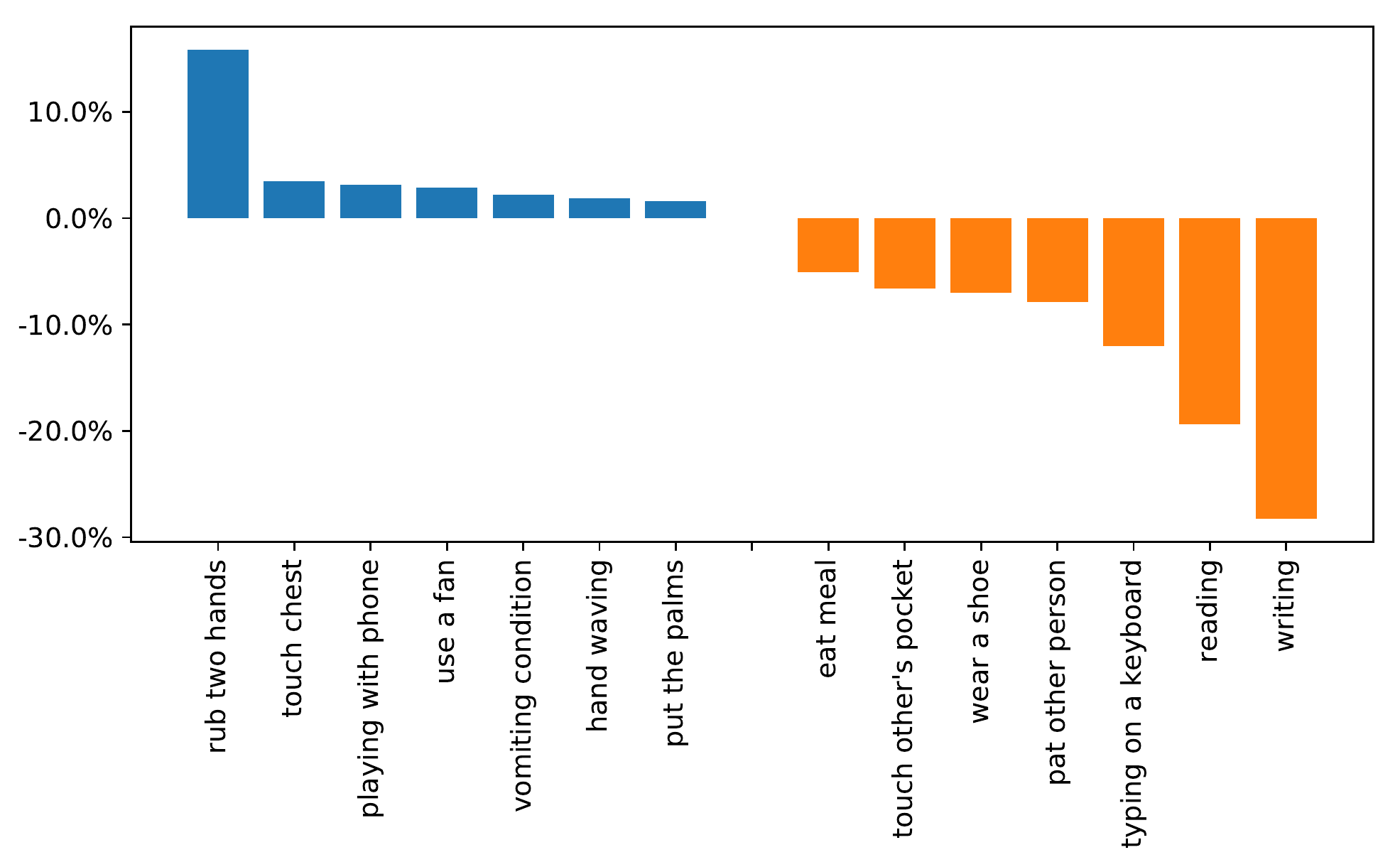}
	\caption{Accuracy difference between the skeleton modality and the RGB modality, i.e., ACC(skeleton)-ACC(RGB).}
	\label{fig:ske-rgb}	
	
	\end{figure}  	
	
    \begin{figure}[!htb]
	\centering
	\includegraphics[width=0.8\linewidth]{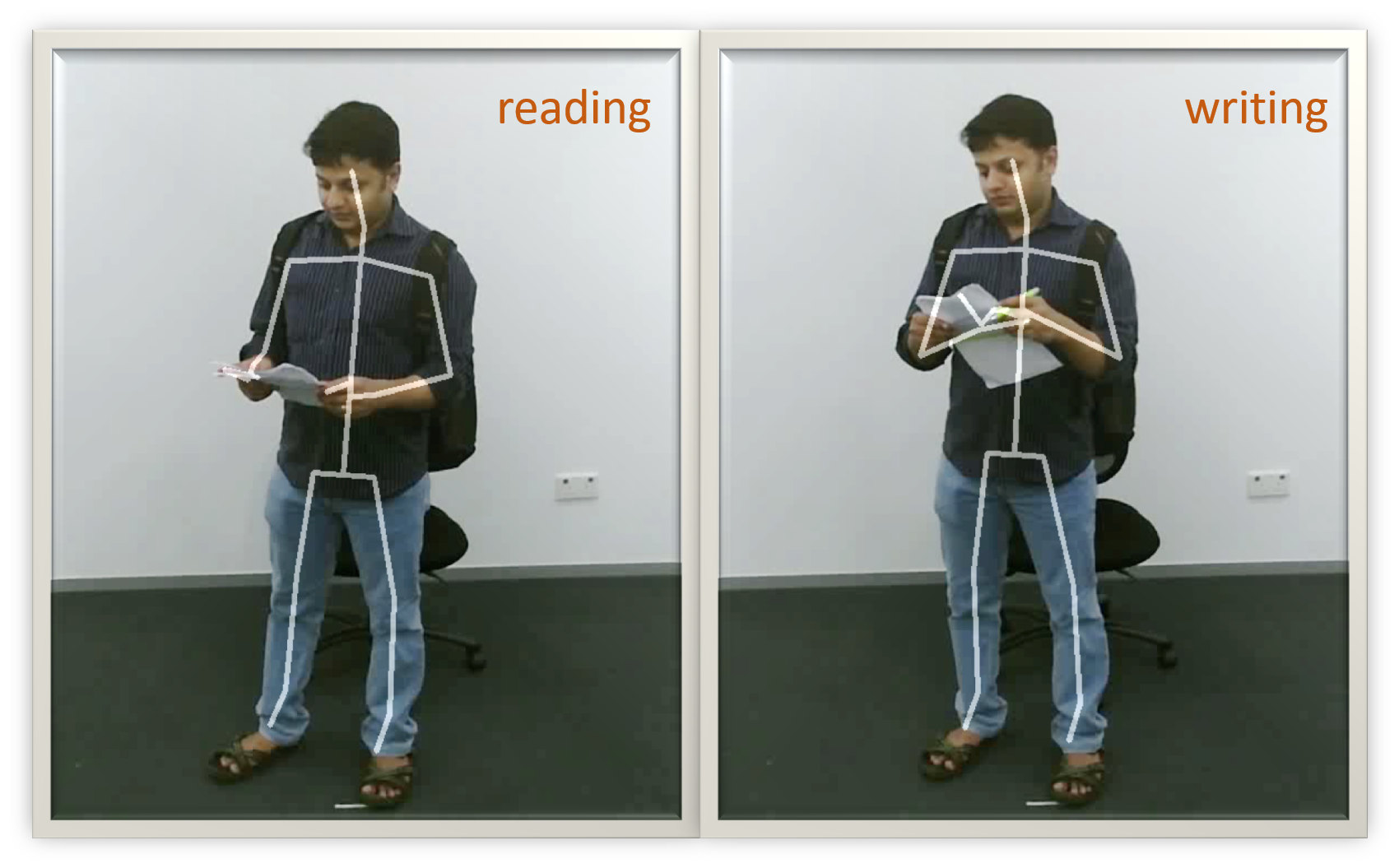}
	\caption{Two examples for class "reading" and class "writing". The white lines represent the skeletons.}
	\label{fig:readvswrite}	
	
	\end{figure}

    Fig.~\ref{fig:ske-rgb} shows the accuracy differences between the skeleton modality and the RGB modality. 
    It shows that the skeleton modality helps RGB modality a lot for "rub the hands" class and the RGB modality helps the skeleton modality a lot for "reading" and "writing" classes. 
    We find two examples for the  class "reading" and class "writing" as shown in Fig.~\ref{fig:readvswrite}. The skeletons of these two examples are very similar, thus are hard to tell apart. 
    But with the help of the RGB data, they can be distinguished according to whether there is a pen in the hands. 
    This example illustrates the complementarity between the skeleton modality and the RGB modality. 

	

		
	

	\subsection{Conclusion}
	\label{sec:conclusion}
    In this work, we propose a novel multi-stream attention-enhanced adaptive graph convolutional neural network (MS-AAGCN) for skeleton-based action recognition. 
    The graph topology of the skeleton data used in this model is parameterized and embedded into the network to be jointly learned and updated with the other parameters. 
    This data-driven approach increases the flexibility and generalization capacity of the model.
    It is also confirmed that the adaptively learned topology of graph is more suitable for the action recognition task than the human-body-based graph. 
    Besides, an STC-attention module is embedded in every graph convolutional layers, which helps the model paying more attention to the important joints, frames and features. 
    Moreover, we propose to explicitly model the joints, bones and the corresponding motion information in a unified multi-stream framework, which further enhances the performance. 
    The final model is evaluated on two large-scale action recognition datasets, i.e., the NTU-RGBD and the Skeleton-Kinetics. 
    It achieves the state-of-the-art performance on both of them. 
    In addition, we fuse the skeleton data with the skeleton-guided cropped RGB data, which brings additional improvement.
    Future works can focus on how to better fuse the RGB modality and the skeleton modality. 
    It is also worth to study the combination of the skeleton-based action recognition algorithms and the pose estimation algorithms in a unified framework.

    
{\small
\bibliographystyle{IEEEtran}
\bibliography{Zotero}

\begin{thebibliography}{10}
\providecommand{\url}[1]{#1}
\csname url@samestyle\endcsname
\providecommand{\newblock}{\relax}
\providecommand{\bibinfo}[2]{#2}
\providecommand{\BIBentrySTDinterwordspacing}{\spaceskip=0pt\relax}
\providecommand{\BIBentryALTinterwordstretchfactor}{4}
\providecommand{\BIBentryALTinterwordspacing}{\spaceskip=\fontdimen2\font plus
\BIBentryALTinterwordstretchfactor\fontdimen3\font minus
  \fontdimen4\font\relax}
\providecommand{\BIBforeignlanguage}[2]{{%
\expandafter\ifx\csname l@#1\endcsname\relax
\typeout{** WARNING: IEEEtran.bst: No hyphenation pattern has been}%
\typeout{** loaded for the language `#1'. Using the pattern for}%
\typeout{** the default language instead.}%
\else
\language=\csname l@#1\endcsname
\fi
#2}}
\providecommand{\BIBdecl}{\relax}
\BIBdecl

\bibitem{wang_action_2013}
H.~Wang and C.~Schmid, ``Action {Recognition} with {Improved} {Trajectories},''
  in \emph{The {IEEE} {International} {Conference} on {Computer} {Vision}
  ({ICCV})}, Dec. 2013, pp. 3551--3558.

\bibitem{simonyan_two-stream_2014}
K.~Simonyan and A.~Zisserman, ``Two-stream convolutional networks for action
  recognition in videos,'' in \emph{Advances in {Neural} {Information}
  {Processing} {Systems}}, 2014, pp. 568--576.

\bibitem{tran_learning_2015}
D.~Tran, L.~Bourdev, R.~Fergus, L.~Torresani, and M.~Paluri, ``Learning
  {Spatiotemporal} {Features} {With} 3d {Convolutional} {Networks},'' in
  \emph{The {IEEE} {International} {Conference} on {Computer} {Vision}
  ({ICCV})}, Dec. 2015, pp. 4489--4497.

\bibitem{carreira_quo_2017}
J.~Carreira and A.~Zisserman, ``Quo {Vadis}, {Action} {Recognition}? {A} {New}
  {Model} and the {Kinetics} {Dataset},'' in \emph{The {IEEE} {Conference} on
  {Computer} {Vision} and {Pattern} {Recognition} ({CVPR})}, Jul. 2017.

\bibitem{wang_temporal_2018}
Y.~Wang, L.~Zhou, and Y.~Qiao, ``Temporal hallucinating for action recognition
  with few still images,'' in \emph{The {IEEE} {Conference} on {Computer}
  {Vision} and {Pattern} {Recognition} ({CVPR})}, 2018, pp. 5314--5322.

\bibitem{vemulapalli_human_2014}
R.~Vemulapalli, F.~Arrate, and R.~Chellappa, ``Human action recognition by
  representing 3d skeletons as points in a lie group,'' in \emph{The {IEEE}
  {Conference} on {Computer} {Vision} and {Pattern} {Recognition} ({CVPR})},
  2014, pp. 588--595.

\bibitem{fernando_modeling_2015}
B.~Fernando, E.~Gavves, J.~M. Oramas, A.~Ghodrati, and T.~Tuytelaars,
  ``Modeling video evolution for action recognition,'' in \emph{The {IEEE}
  {Conference} on {Computer} {Vision} and {Pattern} {Recognition} ({CVPR})},
  2015, pp. 5378--5387.

\bibitem{du_hierarchical_2015}
Y.~Du, W.~Wang, and L.~Wang, ``Hierarchical recurrent neural network for
  skeleton based action recognition,'' in \emph{The {IEEE} {Conference} on
  {Computer} {Vision} and {Pattern} {Recognition} ({CVPR})}, 2015, pp.
  1110--1118.

\bibitem{shahroudy_ntu_2016}
A.~Shahroudy, J.~Liu, T.-T. Ng, and G.~Wang, ``{NTU} {RGB}+{D}: {A} {Large}
  {Scale} {Dataset} for 3d {Human} {Activity} {Analysis},'' in \emph{The {IEEE}
  {Conference} on {Computer} {Vision} and {Pattern} {Recognition} ({CVPR})},
  2016, pp. 1010--1019.

\bibitem{liu_spatio-temporal_2016}
J.~Liu, A.~Shahroudy, D.~Xu, and G.~Wang, ``Spatio-{Temporal} {LSTM} with
  {Trust} {Gates} for 3d {Human} {Action} {Recognition},'' in \emph{The
  {European} {Conference} on {Computer} {Vision} ({ECCV})}, 2016, vol. 9907,
  pp. 816--833.

\bibitem{song_end--end_2017}
S.~Song, C.~Lan, J.~Xing, W.~Zeng, and J.~Liu, ``An {End}-to-{End}
  {Spatio}-{Temporal} {Attention} {Model} for {Human} {Action} {Recognition}
  from {Skeleton} {Data}.'' in \emph{{AAAI}}, 2017, pp. 4263--4270.

\bibitem{zhang_view_2017}
P.~Zhang, C.~Lan, J.~Xing, W.~Zeng, J.~Xue, and N.~Zheng, ``View {Adaptive}
  {Recurrent} {Neural} {Networks} for {High} {Performance} {Human} {Action}
  {Recognition} {From} {Skeleton} {Data},'' in \emph{The {IEEE} {Conference} on
  {Computer} {Vision} and {Pattern} {Recognition} ({CVPR})}, 2017, pp.
  2117--2126.

\bibitem{li_skeleton-based_2018}
L.~Li, W.~Zheng, Z.~Zhang, Y.~Huang, and L.~Wang, ``Skeleton-{Based}
  {Relational} {Modeling} for {Action} {Recognition},''
  \emph{arXiv:1805.02556}, 2018.

\bibitem{li_independently_2018}
S.~Li, W.~Li, C.~Cook, C.~Zhu, and Y.~Gao, ``Independently recurrent neural
  network (indrnn): {Building} {A} longer and deeper {RNN},'' in \emph{The
  {IEEE} {Conference} on {Computer} {Vision} and {Pattern} {Recognition}
  ({CVPR})}, 2018, pp. 5457--5466.

\bibitem{kim_interpretable_2017}
T.~S. Kim and A.~Reiter, ``Interpretable 3d human action analysis with temporal
  convolutional networks,'' in \emph{The {IEEE} {Conference} on {Computer}
  {Vision} and {Pattern} {Recognition} ({CVPR})}, 2017, pp. 1623--1631.

\bibitem{ke_new_2017}
Q.~Ke, M.~Bennamoun, S.~An, F.~A. Sohel, and F.~Boussaïd, ``A {New}
  {Representation} of {Skeleton} {Sequences} for 3d {Action} {Recognition},''
  \emph{IEEE Conference on Computer Vision and Pattern Recognition (CVPR)}, pp.
  4570--4579, 2017.

\bibitem{liu_enhanced_2017}
M.~Liu, H.~Liu, and C.~Chen, ``Enhanced skeleton visualization for view
  invariant human action recognition,'' \emph{Pattern Recognition}, vol.~68,
  pp. 346--362, 2017.

\bibitem{li_skeleton-based_2017}
C.~Li, Q.~Zhong, D.~Xie, and S.~Pu, ``Skeleton-based action recognition with
  convolutional neural networks,'' in \emph{{IEEE} {International} {Conference}
  on {Multimedia} \& {Expo} {Workshops} ({ICMEW})}, 2017, pp. 597--600.

\bibitem{li_skeleton_2017}
B.~Li, Y.~Dai, X.~Cheng, H.~Chen, Y.~Lin, and M.~He, ``Skeleton based action
  recognition using translation-scale invariant image mapping and multi-scale
  deep {CNN},'' in \emph{{IEEE} {International} {Conference} on {Multimedia} \&
  {Expo} {Workshops} ({ICMEW})}, 2017, pp. 601--604.

\bibitem{yan_spatial_2018}
S.~Yan, Y.~Xiong, and D.~Lin, ``Spatial {Temporal} {Graph} {Convolutional}
  {Networks} for {Skeleton}-{Based} {Action} {Recognition},'' in \emph{{AAAI}},
  2018.

\bibitem{tang_deep_2018}
Y.~Tang, Y.~Tian, J.~Lu, P.~Li, and J.~Zhou, ``Deep {Progressive}
  {Reinforcement} {Learning} for {Skeleton}-{Based} {Action} {Recognition},''
  in \emph{The {IEEE} {Conference} on {Computer} {Vision} and {Pattern}
  {Recognition} ({CVPR})}, 2018.

\bibitem{zhang_egogesture:_2018}
Y.~Zhang, C.~Cao, J.~Cheng, and H.~Lu, ``{EgoGesture}: {A} {New} {Dataset} and
  {Benchmark} for {Egocentric} {Hand} {Gesture} {Recognition},'' \emph{IEEE
  Transactions on Multimedia}, pp. 1038--1050, 2018.

\bibitem{kipf_semi-supervised_2016}
T.~N. Kipf and M.~Welling, ``Semi-{Supervised} {Classification} with {Graph}
  {Convolutional} {Networks},'' \emph{arXiv:1609.02907}, Sep. 2016.

\bibitem{duvenaud_convolutional_2015}
D.~K. Duvenaud, D.~Maclaurin, J.~Iparraguirre, R.~Bombarell, T.~Hirzel,
  A.~Aspuru-Guzik, and R.~P. Adams, ``Convolutional {Networks} on {Graphs} for
  {Learning} {Molecular} {Fingerprints},'' in \emph{Advances in {Neural}
  {Information} {Processing} {Systems}}, 2015, pp. 2224--2232.

\bibitem{niepert_learning_2016}
M.~Niepert, M.~Ahmed, and K.~Kutzkov, ``Learning convolutional neural networks
  for graphs,'' in \emph{International {Conference} on {Machine} {Learning}
  ({ICML})}, 2016, pp. 2014--2023.

\bibitem{atwood_diffusion-convolutional_2016}
J.~Atwood and D.~Towsley, ``Diffusion-convolutional neural networks,'' in
  \emph{Advances in {Neural} {Information} {Processing} {Systems}}, 2016, pp.
  1993--2001.

\bibitem{hamilton_inductive_2017}
W.~Hamilton, Z.~Ying, and J.~Leskovec, ``Inductive representation learning on
  large graphs,'' in \emph{Advances in {Neural} {Information} {Processing}
  {Systems}}, 2017, pp. 1025--1035.

\bibitem{monti_geometric_2017}
F.~Monti, D.~Boscaini, J.~Masci, E.~Rodola, J.~Svoboda, and M.~M. Bronstein,
  ``Geometric deep learning on graphs and manifolds using mixture model
  {CNNs},'' in \emph{The {IEEE} {Conference} on {Computer} {Vision} and
  {Pattern} {Recognition} ({CVPR})}, 2017, pp. 5115--5124.

\bibitem{kipf_neural_2018}
T.~Kipf, E.~Fetaya, K.-C. Wang, M.~Welling, and R.~Zemel, ``Neural relational
  inference for interacting systems,'' in \emph{International {Conference} on
  {Machine} {Learning} ({ICML})}, 2018, pp. 2693--2702.

\bibitem{xu_show_2015}
K.~Xu, J.~Ba, R.~Kiros, K.~Cho, A.~C. Courville, R.~Salakhutdinov, R.~S. Zemel,
  and Y.~Bengio, ``Show, {Attend} and {Tell}: {Neural} {Image} {Caption}
  {Generation} with {Visual} {Attention},'' in \emph{International {Conference}
  on {Machine} {Learning} ({ICML})}, 2015, pp. 2048--2057.

\bibitem{wang_residual_2017}
F.~Wang, M.~Jiang, C.~Qian, S.~Yang, C.~Li, H.~Zhang, X.~Wang, and X.~Tang,
  ``Residual {Attention} {Network} for {Image} {Classification},'' in \emph{The
  {IEEE} {Conference} on {Computer} {Vision} and {Pattern} {Recognition}
  ({CVPR})}, Jul. 2017, pp. 6156--3164.

\bibitem{vaswani_attention_2017}
A.~Vaswani, N.~Shazeer, N.~Parmar, J.~Uszkoreit, L.~Jones, A.~N. Gomez,
  Å.~Kaiser, and I.~Polosukhin, ``Attention is {All} you {Need},'' in
  \emph{Advances in {Neural} {Information} {Processing} {Systems}}, 2017, pp.
  6000--6010.

\bibitem{hu_gather-excite:_2018}
J.~Hu, L.~Shen, S.~Albanie, G.~Sun, and A.~Vedaldi, ``Gather-{Excite}:
  {Exploiting} {Feature} {Context} in {Convolutional} {Neural} {Networks},'' in
  \emph{Advances in {Neural} {Information} {Processing} {Systems}}, 2018, pp.
  9423--9433.

\bibitem{kay_kinetics_2017}
W.~Kay, J.~Carreira, K.~Simonyan, B.~Zhang, C.~Hillier, S.~Vijayanarasimhan,
  F.~Viola, T.~Green, T.~Back, P.~Natsev, and {others}, ``The {Kinetics}
  {Human} {Action} {Video} {Dataset},'' \emph{arXiv:1705.06950}, 2017.

\bibitem{shi_two-stream_2019}
L.~Shi, Y.~Zhang, J.~Cheng, and H.~Lu, ``Two-{Stream} {Adaptive} {Graph}
  {Convolutional} {Networks} for {Skeleton}-{Based} {Action} {Recognition},''
  in \emph{The {IEEE} {Conference} on {Computer} {Vision} and {Pattern}
  {Recognition} ({CVPR})}, 2019.

\bibitem{si_skeleton-based_2018}
C.~Si, Y.~Jing, W.~Wang, L.~Wang, and T.~Tan, ``Skeleton-{Based} {Action}
  {Recognition} with {Spatial} {Reasoning} and {Temporal} {Stack} {Learning},''
  in \emph{The {European} {Conference} on {Computer} {Vision} ({ECCV})}, Sep.
  2018.

\bibitem{cao_skeleton-based_2018}
C.~Cao, C.~Lan, Y.~Zhang, W.~Zeng, H.~Lu, and Y.~Zhang, ``Skeleton-{Based}
  {Action} {Recognition} with {Gated} {Convolutional} {Neural} {Networks},''
  \emph{IEEE Transactions on Circuits and Systems for Video Technology}, 2018.

\bibitem{shuman_emerging_2013}
D.~I. Shuman, S.~K. Narang, P.~Frossard, A.~Ortega, and P.~Vandergheynst, ``The
  emerging field of signal processing on graphs: {Extending} high-dimensional
  data analysis to networks and other irregular domains,'' \emph{IEEE Signal
  Processing Magazine}, vol.~30, no.~3, pp. 83--98, 2013.

\bibitem{bruna_spectral_2014}
J.~Bruna, W.~Zaremba, A.~Szlam, and Y.~LeCun, ``Spectral {Networks} and
  {Locally} {Connected} {Networks} on {Graphs},'' in \emph{International
  {Conference} on {Learning} {Representations}}, 2014.

\bibitem{henaff_deep_2015}
M.~Henaff, J.~Bruna, and Y.~LeCun, ``Deep convolutional networks on
  graph-structured data,'' \emph{arXiv:1506.05163}, 2015.

\bibitem{defferrard_convolutional_2016}
M.~Defferrard, X.~Bresson, and P.~Vandergheynst, ``Convolutional {Neural}
  {Networks} on {Graphs} with {Fast} {Localized} {Spectral} {Filtering},'' in
  \emph{Advances in {Neural} {Information} {Processing} {Systems}}, 2016, pp.
  3844--3852.

\bibitem{wang_videos_2018}
X.~Wang and A.~Gupta, ``Videos as {Space}-{Time} {Region} {Graphs},''
  \emph{arXiv:1806.01810}, 2018.

\bibitem{he_deep_2016}
K.~He, X.~Zhang, S.~Ren, and J.~Sun, ``Deep {Residual} {Learning} for {Image}
  {Recognition},'' in \emph{The {IEEE} {Conference} on {Computer} {Vision} and
  {Pattern} {Recognition} ({CVPR})}, Jun. 2016.

\bibitem{cao_realtime_2017}
Z.~Cao, T.~Simon, S.-E. Wei, and Y.~Sheikh, ``Realtime multi-person 2d pose
  estimation using part affinity fields,'' in \emph{The {IEEE} {Conference} on
  {Computer} {Vision} and {Pattern} {Recognition} ({CVPR})}, 2017.

\bibitem{paszke_automatic_2017}
A.~Paszke, S.~Gross, S.~Chintala, G.~Chanan, E.~Yang, Z.~DeVito, Z.~Lin,
  A.~Desmaison, L.~Antiga, and A.~Lerer, ``Automatic differentiation in
  {PyTorch},'' in \emph{Advancesin {Neural} {Information} {Processing}
  {Systems} {Workshops}}, 2017.

\bibitem{shi_gesture_2019}
L.~Shi, Y.~Zhang, C.~Jian, and L.~Hanqing, ``Gesture {Recognition} using
  {Spatiotemporal} {Deformable} {Convolutional} {Represention},'' in
  \emph{{ICIP}}, Oct. 2019.

\bibitem{shahroudy_deep_2018}
A.~Shahroudy, T.~Ng, Y.~Gong, and G.~Wang, ``Deep {Multimodal} {Feature}
  {Analysis} for {Action} {Recognition} in {RGB}+{D} {Videos},'' \emph{IEEE
  Transactions on Pattern Analysis and Machine Intelligence}, vol.~40, no.~5,
  pp. 1045--1058, May 2018.

\bibitem{zolfaghari_chained_2017}
M.~Zolfaghari, G.~L. Oliveira, N.~Sedaghat, and T.~Brox, ``Chained
  {Multi}-{Stream} {Networks} {Exploiting} {Pose}, {Motion}, and {Appearance}
  for {Action} {Classification} and {Detection},'' in \emph{The {IEEE}
  {International} {Conference} on {Computer} {Vision} ({ICCV})}, 2017, pp.
  2904--2913.

\bibitem{luvizon_2d/3d_2018}
D.~C. Luvizon, D.~Picard, and H.~Tabia, ``2d/3d {Pose} {Estimation} and
  {Action} {Recognition} {Using} {Multitask} {Deep} {Learning},'' in \emph{The
  {IEEE} {Conference} on {Computer} {Vision} and {Pattern} {Recognition}
  ({CVPR})}, 2018, pp. 5137--5146.

\bibitem{baradel_glimpse_2018}
F.~Baradel, C.~Wolf, J.~Mille, and G.~W. Taylor, ``Glimpse {Clouds}: {Human}
  {Activity} {Recognition} {From} {Unstructured} {Feature} {Points},'' in
  \emph{The {IEEE} {Conference} on {Computer} {Vision} and {Pattern}
  {Recognition} ({CVPR})}, 2018, pp. 469--478.

\bibitem{liu_recognizing_2018}
M.~Liu and J.~Yuan, ``Recognizing {Human} {Actions} as the {Evolution} of
  {Pose} {Estimation} {Maps},'' in \emph{The {IEEE} {Conference} on {Computer}
  {Vision} and {Pattern} {Recognition} ({CVPR})}, 2018, pp. 1159--1168.

\bibitem{das_where_2019}
S.~Das, A.~Chaudhary, F.~Bremond, and M.~Thonnat, ``Where to {Focus} on for
  {Human} {Action} {Recognition}?'' in \emph{{IEEE} {Winter} {Conference} on
  {Applications} of {Computer} {Vision} ({WACV})}, Jan. 2019, pp. 71--80.

\end{thebibliography}
}

\end{document}